\documentclass[A4,11pt,oneside]{article}
\usepackage{hyperref}
\hypersetup{
    colorlinks=true,
    linkcolor=blue,
    filecolor=magenta,      
    urlcolor=cyan,
}
\usepackage{amsmath}
\usepackage{amsfonts}
\usepackage{amssymb}
\usepackage{stmaryrd}
\usepackage[dvipsnames]{pstricks}

\usepackage{graphicx}
\usepackage{subfig}

\usepackage{hhline}
\graphicspath{{eps/}}
\DeclareMathOperator*{\argmin}{argmin}
\newcommand{\R}{\mathbb{R}}

\def\TR#1{\textcolor{black}{#1}}

\DeclareMathOperator{\T}{Tr}
\DeclareMathOperator{\Hess}{Hess}

\usepackage{geometry}
\usepackage{caption}
\begin{document}

\title{Efficient Seismic fragility curve estimation by Active Learning on Support Vector Machines}

\author{
   R\'emi Sainct\thanks{Den-Service d'études mécaniques et thermiques (SEMT), CEA, Université Paris-Saclay, F-91191 Gif-sur-Yvette, France. Email: \href{mailto:remi.sainct@m4x.org}{remi.sainct@m4x.org}}
   \and
   Cyril Feau\thanks{Den-Service d'études mécaniques et thermiques (SEMT), CEA, Université Paris-Saclay, F-91191 Gif-sur-Yvette, France. Email: \href{mailto:cyril.feau@cea.fr}{cyril.feau@cea.fr}}
   \and
   Jean-Marc Martinez\thanks{Den-Service de thermo-hydraulique et de mécanique des fluides (STMF), CEA, Université Paris-Saclay, F-91191 Gif-sur-Yvette, France. Email: \href{mailto:jean-marc.martinez@cea.fr}{jean-marc.martinez@cea.fr}}
   \and
   Josselin Garnier\thanks{Ecole Polytechnique, CMAP, 91128 Palaiseau Cedex, France. Email:\href{mailto:josselin.garnier@polytechnique.edu}{josselin.garnier@polytechnique.edu}}
} 

\maketitle

\abstract{Fragility curves which express the failure probability of a structure, or critical components, as function of a loading intensity measure are nowadays widely used (i) in Seismic Probabilistic Risk Assessment studies, (ii) to evaluate impact of construction details on the structural performance of installations under seismic excitations or under other loading sources such as wind. To avoid the use of parametric models such as lognormal model to estimate fragility curves from a reduced number of numerical calculations, a methodology based on Support Vector Machines coupled with an active learning algorithm is proposed in this paper. In practice, input excitation is reduced to some relevant parameters and, given these parameters, SVMs are used for a binary classification of the structural responses relative to a limit threshold of exceedance. Since the output is not only binary, this is a score, a probabilistic interpretation of the output is exploited to estimate very efficiently fragility curves as score functions or as functions of classical seismic intensity measures.\\
}

Keywords : Fragility curve, Active Learning, Support Vector Machines, seismic intensity measure indicator.

\section{Introduction}

In the Seismic Probabilistic Risk Assessment (SPRA) studies performed on industrial facilities, a key point is the evaluation of fragility curves which express the failure probability of a structure, or critical components, as a function of a seismic intensity measure such as peak ground acceleration (PGA) or spectral acceleration. It should be noted that apart from the use in a SPRA framework, fragility curves can be used for making decisions regarding the choice of construction details, to improve  the structural performance of installations under seismic excitations (see e.g. \cite{ZHANG20091648, SAHA201620, PATIL201692, doi:10.1002/eqe.2586}). They can also be used to evaluate impact of ground motion characteristics (near-fault type like, broadband, etc.) on the structural vulnerability (see e.g. \cite{doi:10.1002/eqe.2586, WangF2017}). Finally, it is worth noting that the use of fragility curves is not limited to seismic excitation, they can also be applied to other loading sources such as wind for example \cite{QUILLIGAN2012270}.

In theory, for complex structures, fragility curves have to be evaluated empirically based on a large number of mechanical analyses requiring, in most cases, nonlinear time-history calculations including both the uncertainties inherent to the system capacity and to the seismic demand, respectively called epistemic and aleatory uncertainties \cite{KIUREGHIAN2009105}. Nevertheless, the prohibitive computational cost induced by most of nonlinear mechanical models requires the development of numerically efficient methods to evaluate such curves from a minimal number of computations, in particular in industrial contexts.

Following the idea proposed in the early 1980's in the framework of nuclear safety assessment \cite{KENNEDY1980315}, the lognormal parametric model has been widely used in many applications to estimate fragility curves from a reduced number of numerical calculations (see e.g. \cite{ZHANG20091648, SAHA201620, PATIL201692, WangF2017}). Different methods can be used to determine the parameters of the lognormal model (see e.g. \cite{ZENTNER20101614, Baker2015}), however, the question of the validity of this assumption arises. Typically, in \cite{Mai2017} authors show that for a given structure the accuracy of the lognormal curves depends on the ground motion intensity measure, the failure criterion and the employed method for fitting the model. As shown in \cite{doi:10.1002/eqe.2567}, the class of structures considered may also have an influence on the adequacy of the lognormal model.

The question of the representativity is inevitable with the use of parametric models since, for the complex cases of interest, it is very difficult to verify their validity. To bypass this problem, the need of a numerically efficient non-parametric-based methodology (which would be accurate with a minimum number of mechanical analyses) is necessary. A way to achieve this goal consists in building a metamodel (i.e. a surrogate model of the mechanical analysis) which expresses the statistical relation between
seismic inputs and structural outputs. Various metamodeling strategies have been proposed recently in the literature based on, for example, response surfaces (\cite{PARK20141, SEO2013642}), kriging \cite{doi:10.1002/eqe.2586} and Artificial Neural Networks (ANNs) \cite{WANG2018213}.

The goal of this paper is twofold. First, it is to propose a simple and efficient methodology for estimating non-parametric fragility curves that allows to reduce the cost of mechanical numerical computations by optimizing their selection. Second, it is to adress the question of the best seismic intensity measure indicator that can to be used as abscissa of the fragilty curves and not be limited to the PGA. To this end, the strategy proposed is based on the use of Support Vector Machines (SVMs) coupled with an active learning algorithm. Thus, input excitation is reduced to some relevant parameters and, given these parameters, SVMs are used for a binary classification of the structural responses relative to a limit threshold of exceedance. It is worth noting that their output is not only binary, it is a "score" that can have a probabilistic interpretation as we will see.

In contrast to classical learning (passive learning), the active learner selects the most useful numerical experiments to be added to the learning data set. The "learners" choose the best instances from a given large set of unlabeled examples. So, the main question in active learning is how to choose new numerical experiments to be labeled. Various methods proposed in active learning by ANNs are presented in \cite{Hasenjager2002}. Most are based on the learning of several "learners" (\cite{Seung:1992:QC:130385.130417, Gazut:2008:TOD:2325839.2328058}). With SVMs, active learning can be done very easily by using only one learner because the distance to the separator hyperplane is a "natural" criterion for selecting new points to "label" \cite{Ton}. A similar technique using logistic output neural networks can be used by analyzing the logit of the output. But in this case, given the non-linearity of the ANNs, the different learnings of the learner may present a strong variability on the decision boundary.

Finally, as the SVM output is a score, this score can be used as abscissa of the fragility curves. Indeed, a perfect classifier, if it exists, would lead to a fragility curve in the form of a unit step function, i.e. corresponding to a fragility curve "without uncertainty". Nevertheless although certainly not perfect, in the linear binary classification this score can particularly be relevant for engineering purpose since it is a simple linear combination of the input parameters.

To illustrate the proposed methodology, inputs parameters are defined from a set of real accelerograms which is enriched by synthetic accelerograms using the procedure defined in \cite{doi:10.1002/eqe.997}, which is based on a parameterized stochastic model of modulated and filtered white-noise process. A brief summary of this model is presented in section 2 of this paper. Moreover a simple inelastic oscillator is considered in order to validate the methodology at a large scale within a Monte Carlo-based approach that does not require any assumption to estimate probabilities of interest. This physical model is presented in section 3. Section 4 is devoted to the presentation of the different classification methods, and the active learning methodology. Section 5 shows how the proposed methodology makes it possible to estimate fragility curves, using either the score functions or classical intensity measure indicators such as the PGA. Finally, a conclusion is presented in section 6.

\section{Model of earthquake ground motion}

\subsection{Formulation of the model}

Following Rezaeian \cite{Rez}, a seismic ground motion $s(t)$ with $t \in [0,T]$ is modeled as:

\begin{equation} 
s(t) = q(t, \pmb{\alpha} ) \left[ \frac{1}{\sigma_f(t)} \int_{-\infty}^t h[t-\tau,\pmb{\lambda}(\tau)] w(\tau) d\tau \right], \label{eq1}
\end{equation}
where $q(t, \pmb{\alpha} )$ is a deterministic, non-negative modulating function with a set of parameters $\pmb{\alpha}$, and the process inside the squared brackets is a filtered white-noise process of unit variance: $w(t)$ is a white-noice process, $h(t,\pmb{\lambda})$ denotes the impulse response function (IRF) of the linear filter with a set of parameters $\pmb{\lambda}$, and $\sigma_f(t) = \sqrt{\int_{-\infty}^t h^2(t-\tau,\pmb{\lambda}(\tau)) d\tau}$ is the standard deviation of the process defined by the integral in equation \ref{eq1}.

In order to achieve spectral nonstationarity of the ground motion, the parameters $\pmb{\lambda}$ of the filter depend on the time $\tau$ of application of the pulse; thus the standard deviation $\sigma$ depend on $t$. Still following Rezaeian, we choose for the impulse response function:

\begin{equation} \begin{array}{rcl} h[t-\tau, \pmb{\lambda}(\tau)] &=& \displaystyle \frac{\omega_f(\tau)}{\sqrt{1-\zeta_f^2}} \exp \left[ -\zeta_f \omega_f(\tau) (t-\tau) \right]  \sin \left[ \omega_f(\tau) \sqrt{1-\zeta_f^2} (t-\tau) \right] \quad \text{ if } t \geq \tau, \\
&=& 0 \quad \text{ otherwise}, \end{array} \end{equation}
where  $\pmb{\lambda}(\tau) = [ \omega_f(\tau), \zeta_f]$ is the set of parameters, $\omega_f(\tau)$ is the natural frequency (dependent on the time of application of the pulse) and $\zeta_f \in [0,1]$ is the (constant) damping ratio. A linear form is chosen for the  frequency: $\omega_f(\tau) = \omega_0 + \frac{\tau}{T}(\omega_n-\omega_0)$. The modulating function $q(t, \pmb{\alpha} )$ is defined piecewisely:

\begin{equation} \begin{array}{rclcl} q(t,\pmb{\alpha}) &=& 0 &\text{if } & t \leq T_0,\\
&=& \alpha_1 \left( \frac{t-T_0}{T_1-T_0} \right)^2 &\text{if } & T_0 \leq t \leq T_1,\\
&=& \alpha_1 &\text{if } & T_1 \leq t \leq T_2,\\
&=& \alpha_1 \exp \left[ -\alpha_2 (t-T_2)^{\alpha_3} \right] & \text{if } & t \geq T_2. \end{array} 
\end{equation}
The modulation parameters are thus: $\pmb{\alpha} = (\alpha_1, \alpha_2, \alpha_3, T_0, T_1, T_2)$. The initial delay $T_0$ is used in parameter identification for real ground motions, but it is not used in simulations (we choose $T_0=0$). To summarize, the generated signals are associated with $8$ real parameters: $(\alpha_1, \alpha_2, \alpha_3, T_1, T_2, \omega_0, \omega_n, \zeta_f)$. Finally, a high-pass filter is used as post-processing to guarantee zero residuals in the acceleration, velocity and displacement. The corrected signal $\ddot{u}(t)$ is the solution of the differential equation:

\begin{equation} \ddot{u}(t) + 2 \omega_c \dot{u}(t) + \omega_c^2 u(t) = s(t), 
\end{equation}
where $\omega_c = 0.2$ Hz is the corner frequency. Due to high damping of the oscillator (damping ratio of $100\%$), it is clear that $u(t)$, $\dot{u}(t)$ and $\ddot{u}(t)$ all vanish shortly after the input process $s(t)$ has vanished, thus assuring zero residuals for the simulated ground motion. In the rest of this paper we will use $s(t)$ for the corrected signal $\ddot{u}(t)$.

\subsection{Parameter identification}

The first step to generate artificial signals is to identify the 9 model parameters for every real signal $a(t)$ from our given database of $N_r = 97$ acceleration records (selected from the European Strong Motion Database \cite{ESMD} for
$ 5.5 < M < 6.5$ and $R < 20 {\rm km}$, where $M$ is the magnitude and $R$ the distance from the epicenter). Following Rezaeian \cite{Rez,doi:10.1002/eqe.997}, the modulation parameters $\pmb{\alpha}$ and the filter parameters $\pmb{\lambda}$ are identified independently as follows.

\subsubsection{Modulation parameters}
For a target recorded accelerogram $a(t)$, we determine the modulation parameters $\pmb{\alpha}$ by matching the cumulative energy of the accelerogram $E_a(t) = \int_0^t a^2(\tau) d\tau$ with the expected cumulative energy $E_s(t)$ of the stochastic process $s(t)$, which does not depend on the filter parameters (if the high-pass postprocessing is neglected):

\begin{equation} \begin{array}{rll}  E_s(t) & = & \mathbb{E} \left[ {\displaystyle \int_0^t}  s^2(\tau) d\tau \right]  \\
& = & \mathbb{E} \left[ {\displaystyle \int_0^t} q^2(\tau, \pmb{\alpha}) \frac{1}{\sigma_f^2(\tau)} \left({\displaystyle  \int_{-\infty}^{\tau}} h[\tau-u,\pmb{\lambda}(u)] w(u) du \right)^2 d\tau \right] \\
 & = & {\displaystyle \int_0^t} q^2(\tau, \pmb{\alpha}) \frac{1}{\sigma_f^2(\tau)} \mathbb{E} \left[ \left({\displaystyle  \int_{-\infty}^{\tau}} h[\tau-u,\pmb{\lambda}(u)] w(u) du \right)^2 \right] d\tau \\
& = & {\displaystyle \int_0^t} q^2(\tau, \pmb{\alpha}) d\tau. \end{array} \label{energyconservation}\end{equation}

Thanks to the definition of $\sigma_f(t)$, this expected energy only depends on the modulation parameters $\pmb{\alpha}$. To match the two cumulative energy terms, we minimize the integrated squared difference between them:

\begin{equation} \hat{\pmb{\alpha}} = \argmin_{\pmb{\alpha}} \int_0^T \left[ E_s(t) - E_a(t) \right]^2 dt = \argmin_{\pmb{\alpha}} \int_0^T \left[ \int_0^t q^2(\tau, \pmb{\alpha}) d\tau - \int_0^t a^2(\tau)d\tau \right]^2 dt. \end{equation}

This minimization is done with the Matlab function fminunc. Note that the PGA of the generated signal is not necessarily equal to that of the recorded one on average.

\subsubsection{Filter parameters}
For the filter parameters $\pmb{\lambda} = (\omega_0, \omega_n, \zeta_f)$, we use the zero-level up-crossings, and the positive minima and negative maxima of the simulated signal $s(t)$ and target signal $a(t)$. These quantities do not depend on scaling, thus we use only the un-modulated process 

\begin{equation} y(t) = \int_{-\infty}^t \frac{h[t-\tau,\pmb{\lambda}(\tau)]}{\sigma_f(t)} w(\tau) d\tau. \end{equation}

For a given damping ratio $\zeta_f$, we can identify the frequencies $(\omega_0, \omega_n)$ by matching the cumulative count $N_a(t)$ of zero-level up-crossings of the target signal $a(t)$ with the same expected cumulative count $N_x(t)$ for the simulated signal, given by:

\begin{equation} N_x(t) = \int_0^t \nu(\tau) r(\tau) d\tau, 
\end{equation}
where $\nu(\tau)$ is the mean zero-level up-crossing rate of the process $y(t)$ and $r(\tau)$ is an adjustment factor due to discretization (usually between $0.75$ and $1$). Since $y(t)$ is a Gaussian process with zero mean and unit variance, the mean rate $\nu(\tau)$, after simplification, is given by:

\begin{equation} \nu(t) = \frac{\sigma_{\dot{y}}(t)}{2\pi}, 
\end{equation}
where $\sigma_{\dot{y}}(t)$ is the standard deviation of the time derivative $\dot{y}(t)$ of the process:

\begin{equation}
\sigma_{\dot{y}}(t)^2 = 
\int_{-\infty}^t \left[ \dot{h}(t-\tau, \pmb{\lambda}(\tau)) - h(t-\tau, \pmb{\lambda}(\tau))
\frac{\int_{-\infty}^t h(t-u, \pmb{\lambda}(u))\dot{h}(t-u, \pmb{\lambda}(u))du}{\sigma_f(t)^2} \right]^2  
\frac{d\tau}{\sigma_f(t)^2},
\end{equation}
assuming we neglect integrals over a fraction of a time step in the discretization. 

To identify the damping ratio $\zeta_f$, we use the cumulative count of positive minima and negative maxima. Indeed, in a narrow-band process ($\zeta_f$ close to $0$), almost all maxima are positive and almost all minima are negative, but the rate increases with increasing bandwidth (larger $\zeta_f$). An explicit formulation exists for this rate but it involves computing the second derivative of $y(t)$ \cite{adler}, thus it is easier to use a simulation approach, by counting and averaging the negative maxima and positive minima in a sample of simulated realizations of the process, then choosing the value that minimizes the difference between real and simulated cumulative counts.

\subsection{Simulation of ground motions}

So far, we have defined a model of earthquake ground motions, and explained how to identify each of the parameters from a single real signal $a(t)$. The model then allows one to generate any number of artificial signals, thanks to the white noise $w(t)$. However, these signals would all have very similar features; in order to estimate a fragility curve, we need to be able to generate artificial signals over a whole range of magnitudes, with realistic associated probabilities. Thus, we have to add a second level of randomness in the generation process, coming from the parameters themselves. With Rezaian's method we identified all the model parameters $\theta = (\pmb{\alpha},\pmb{\lambda})$ for each of the $N_r = 97$ acceleration records, giving us $N_r$ data points in the parameter space ($\R^8$ in this case). Then, to define the parameters' distribution, we use a Gaussian kernel density estimation (KDE) with a multivariate bandwidth estimation, following Kristan \cite{Kri}.

Let $(\theta_1,\theta_2,\dots,\theta_{N_r})$ be a multivariate independent and identically distributed sample drawn from some distribution with an unknown density $p(\theta)$, $\theta \in \R^d$. The kernel density estimator $p_{KDE}$ is:

\begin{equation} p_{KDE}(\theta) = \frac{1}{N_r} \sum_1^{N_r} \phi_\mathbf{H}(\theta-\theta_i), 
\end{equation}
where $\phi_\mathbf{H}$ is a Gaussian kernel centered at $0$ with covariance matrix $\mathbf{H}$. A classical measure of closeness of the estimator $p_{KDE}(\theta)$ to the unknown underlying probability density function (pdf) is the \textit{asymptotic mean integrated squared error} (AMISE), defined as:

\begin{equation} {\rm AMISE}(\mathbf{H}) = (4\pi)^{-d/2} \vert \mathbf{H} \vert^{-1/2} N_r^{-1} + \frac{1}{4} d^2 \int \T^2 \left[ \mathbf{H} \Hess_p (\theta) \right] d\theta,\end{equation}
where $\T$ is the trace operator and $\Hess_p$ is the Hessian of $p$. If we write $\mathbf{H} = \beta^2 \mathbf{F}$ with $\beta \in \mathbb{R}$ and we suppose that $\mathbf{F}$ is known, then the AMISE is minimized for:

\begin{equation} \beta_{opt} = \left[ d(4\pi)^{d/2}N_r R(p,\mathbf{F}) \right]^{-\frac{1}{d+4}}, \end{equation}
where $R$ still depends on the underlying (and unknown) distribution $p(\theta)$:

\begin{equation} R(p,\mathbf{F}) = \int \T^2\left[ \mathbf{F} \Hess_p(\theta)\right] d\theta. \end{equation}

Following Kristan \cite{Kri}, ${\bf F}$ can be approximated by the empirical covariance matrix $\hat{\boldsymbol{\Sigma}}^{smp }$ of the observed samples and $R(p,{\bf F})$ can be approximated by
\begin{equation}
\hat{R} = 
\Big( \frac{4}{(d+2) N_r}\Big)^{-\frac{4}{d+4}}
\sum_{i,j=1}^{N_r} \phi_{\hat{\bf G}} ( \theta_i-\theta_j )
\Big( \frac{2}{N_r} (1-2m_{ij}) + (1-m_{ij})^2 \Big) ,
\end{equation}
where 
\begin{equation}
m_{ij} = (\theta_i-\theta_j )^T \hat{\bf G}^{-1} (\theta_i-\theta_j ),
\quad \quad 
\hat{\bf G} = 
\Big( \frac{4}{(d+2) N_r}\Big)^{\frac{2}{d+4}}
\hat{\boldsymbol{\Sigma}}^{smp } .
\end{equation}

%We approximate $R$ with
%
%\begin{equation} \hat{R}(p, \mathbf{F}, \mathbf{G}) = \int \T \left[ \mathbf{F} \Hess_{p_\mathbf{G}}(\theta) \right] \T \left[ \mathbf{F} \Hess_{p_s}(\theta) \right] d\theta, \label{hatr} \end{equation}
%
%using, on the one hand, the empirical distribution $p_s$ given by the samples, and on the other hand a \textit{pilot distribution} $p_\mathbf{G}$, defined as:
%
%\begin{equation} p_\mathbf{G}(\theta) = \phi_\mathbf{G} * p_s(\theta) = \frac{1}{N_r} \sum_1^{N_r} \phi_\mathbf{G}(\theta-\theta_i). \end{equation}
%
%Since $p_s(\theta)$ and $p_\mathbf{G}(\theta)$ are both Gaussian mixture models, we can calculate $\hat{R}$ using only matrix algebra. Finally, the structure $\mathbf{F}$ and the pilot bandwidth $\mathbf{G}$ of the distribution $p_\mathbf{G}$ are approximated using the empirical covariance $\hat{\Sigma}_{smp}$ of the observed samples:
%
%\begin{equation} \mathbf{F} = \hat{\Sigma}_{smp}, \end{equation}
%\begin{equation} \mathbf{G} = \hat{\Sigma}_{smp} \left( \frac{4}{(d+2)N_r} \right)^{\frac{2}{d+4}}, \end{equation}
%
%where the scale of the bandwith $\mathbf{G}$ is estimated by a multivariate Gaussian approximation, i.e. Silverman's rule of thumb.\\

The estimator $p_{KDE}(\theta) = \frac{1}{N_r} \sum_1^{N_r} \phi_\mathbf{H}(\theta-\theta_i)$ is now fully defined. The simulation of an artificial ground motion thus requires three steps:\begin{itemize}

\item choose an integer $i \in \llbracket 1,N_r\rrbracket$ with a uniform distribution;
\item sample a vector $\mathbf{y}$ from a Gaussian distribution with pdf $\phi_\mathbf{H}$ centered at $0$ with covariance matrix $\mathbf{H}$, and let $\theta = \theta_i + \mathbf{y}$;
\item sample a white noise $w(\tau)$ and compute the signal using (\ref{eq1}), with parameters $(\pmb{\alpha},\pmb{\lambda}) = \theta$.
\end{itemize}

\noindent For this article we generated $N_s = 10^5$ artificial seismic ground motions $s_i(t)$ using this method.

\section{Physical model}

\subsection{Equations of motion}

For the illustrative application of the methodology developed in this paper, a nonlinear single degree of freedom system is considered. Indeed, despite its extreme simplicity, such model may reflect the essential features of the nonlinear responses of some real structures. Moreover, in a probabilistic context requiring Monte Carlo simulations, it makes it possible to have reference results with reasonable numerical cost. Its equation of motion reads:

\begin{equation}
\ddot{z}_i(t) + 2 \beta \omega_L \dot{z}_i(t) + f^{nl}_i(t) = - s_i(t), \quad i \in \llbracket 1,N_s \rrbracket
\label{PhysMod}
\end{equation}
where $\dot{z}_i(t)$ and $\ddot{z}_i(t)$ are respectively the relative velocity and acceleration of the unit mass of the system submitted to the ith artificial seismic ground motion $s_i(t)$ with null initial conditions in velocity and displacement. In equation \ref{PhysMod}, $\beta$ is the damping ratio, $\omega_L = 2 \pi f_L$ is the circular frequency and $ f^{nl}_i(t)$ is the nonlinear resisting force. In this study, $f_L = 5$ Hz, $\beta = 2\%$, the yield limit is $Y = 5.10^{-3}$ m, and the post-yield stiffness, defining kinematic hardening, is equal to $20 \%$ of the elastic stiffness. Moreover, we call $\tilde{z}_i(t)$ the relative displacement of the associated linear system, that is assumed to be known in the sequel, whose equation of motion is:

\begin{equation}
\ddot{\tilde{z}}_i(t) + 2 \beta \omega_L \dot{\tilde{z}}_i(t) + \omega_L^2 \tilde{z}_i(t) = - s_i(t),
\label{PhysMod_L}
\end{equation}
and we set: 
\begin{equation}
Z_i = \text{max}_{t\in[0,T]}|z_i(t)|,
\label{Def_Zi}
\end{equation} 
 
\begin{equation}
L_i = \text{max}_{t\in[0,T]}|\tilde{z}_i(t)|.
\label{Def_Li}
\end{equation}

In this work, equations \ref{PhysMod} and \ref{PhysMod_L} are solved numerically with a finite-difference method.

\subsection{Response spectrum}

\begin{figure}[!ht]
\centering
\subfloat[\label{SROa}]{\includegraphics[width=6.7cm]{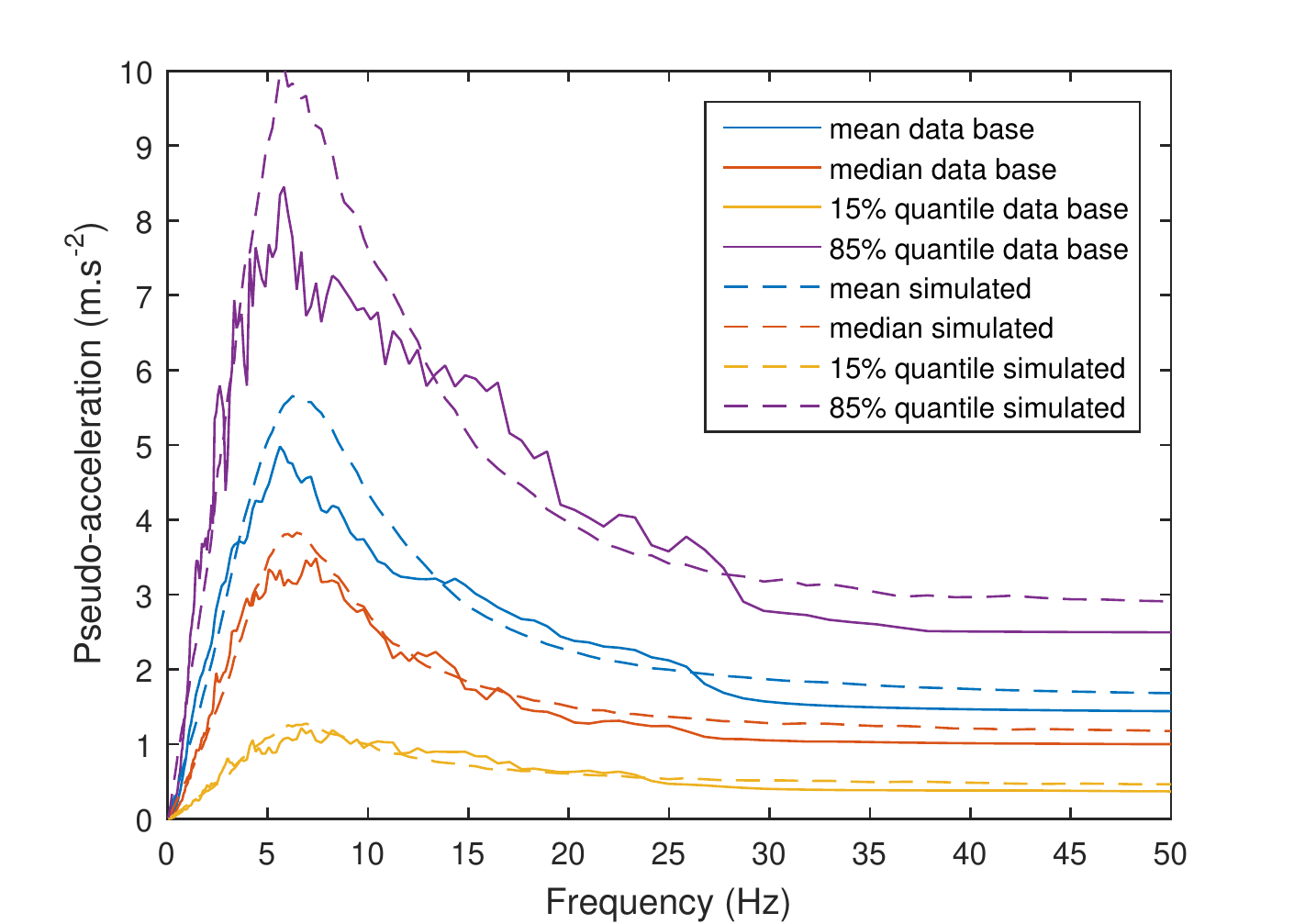}}\\
\subfloat[\label{SROb}]{\includegraphics[width=5.7cm]{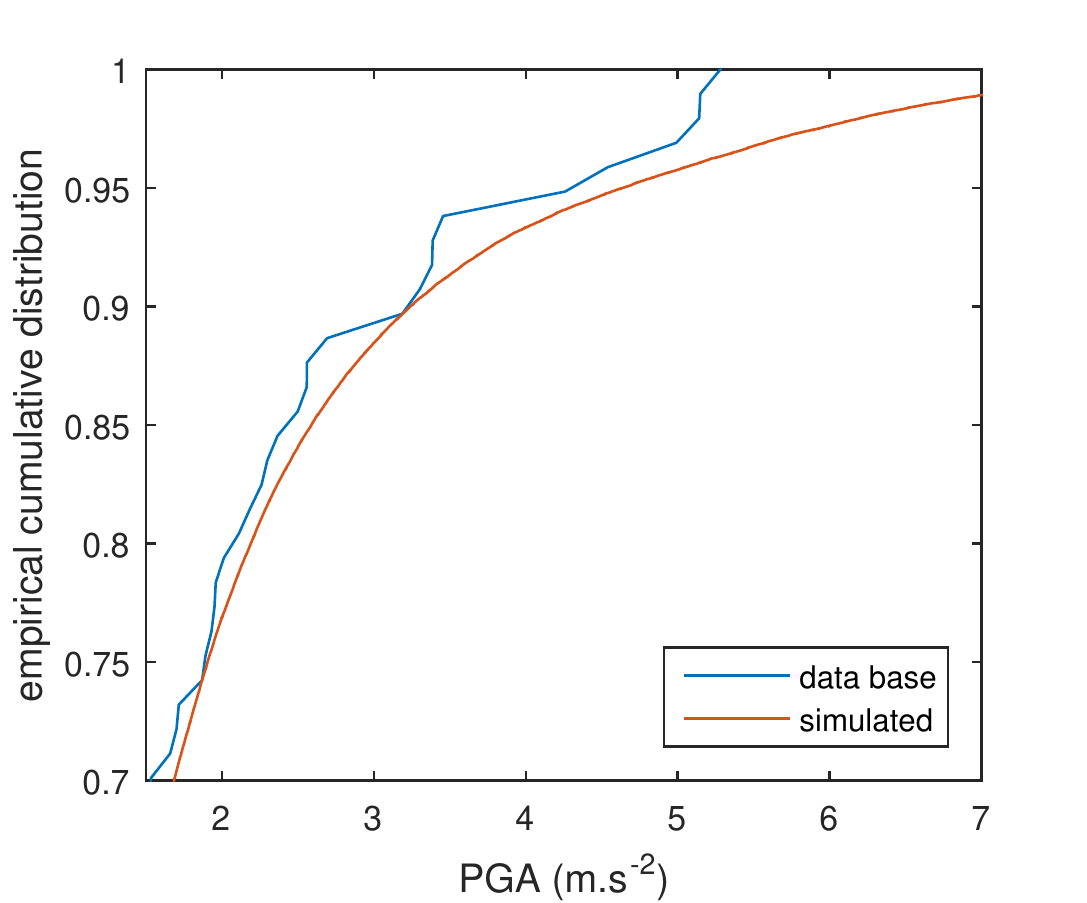}}
\subfloat[\label{SROc}]{\includegraphics[width=5.7cm]{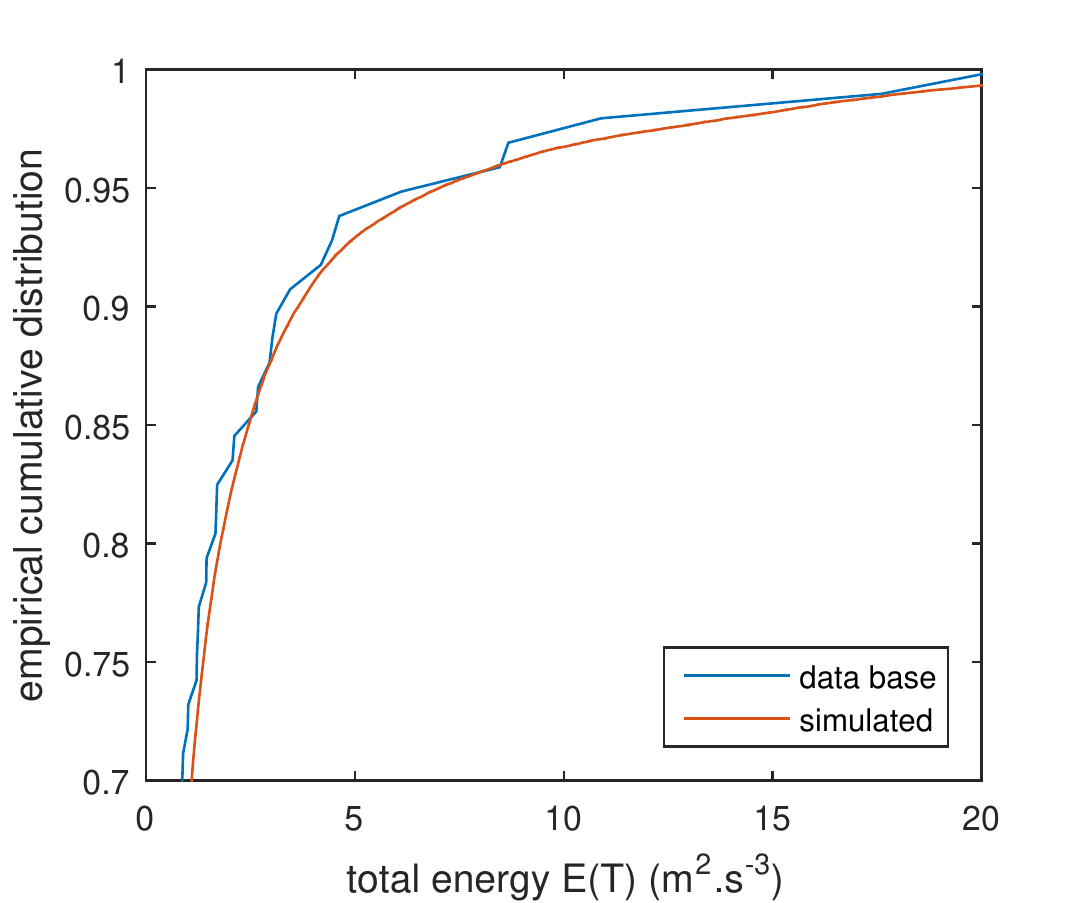}}
\caption{Comparison between the real and simulated data bases. (a) Response spectra for $2\%$ damping ratio. Zoom on the empirical cumulative distribution functions of (b) the PGA and (c) total energy.}
\label{SRO}
\end{figure}

Using the linear equation \ref{PhysMod_L}, we can compare the response spectra of the $N_r = 97$ recorded accelerograms with that of the $N_s = 10^5$ simulated signals. Figure \ref{SROa} shows this comparison for the average spectrum, as well as the $0.15$, $0.5$ and $0.85$ quantiles. It can be seen that the simulated signals have statistically the same response spectra than the real signals, although at high frequency ($f_L > 30$ Hz), the strongest simulated signals have higher responses than the real ones. This may be due to the fact that the model conserves energy (equation \ref{energyconservation}), while the selection of acceleration records from ESMD is based on magnitude. 
%As a consequence, the highest PGA among the simulated signals (12 $\text{m.s}^{-2}$) is significantly higher that the highest PGA of the recorded ones (5.3 $\text{m.s}^{-2}$). 
To illustrate this, figures \ref{SROb} and \ref{SROc} show the empirical cumulative distribution functions of the PGA and total energy. While there is a good match for the energy, the PGA of the strongest simulated signals is slightly higher than for the real ones.

\subsection{Choice of the seismic intensity measures}

Let $\mathcal{B} = (s_i(t))_{i\in \llbracket 1,N_s \rrbracket}$ be our database of simulated ground motions, and $\theta_i = (\pmb{\alpha}_i,\pmb{\lambda}_i) \in \R^8$ the associated modulating and filter parameters. For every signal $s_i(t)$, we also consider: \begin{itemize}
\item the peak ground acceleration $PGA_i = \max_{t\in [0,T]} \left\vert s_i(t) \right\vert$;
\item the maximum velocity (or Peak Ground Velocity) $V_i = \max_{t\in [0,T]} \left\vert \int_0^t s_i(\tau) d\tau \right\vert $;
\item the maximum displacement (or Peak Ground Displacement) $D_i = \max_{t\in [0,T]} \left\vert \int_0^t \int_0^\tau s_i(u) du d\tau \right\vert $;
\item the total energy $E_i = E_{s_i}(T) = \int_0^T s_i^2(\tau) d\tau$;
\item the maximum linear displacement $L_i$ of the structure (equation \ref{Def_Li}). \TR{Conventionnaly, this is spectral acceleration ($\omega_L^2 L_i$) which is considered as intensity measure indicator. Nevertheless, since here the variable of interest is a non-linear displacement, it is more suitable to use spectral displacement.}
%since this value is much faster to compute than the maximum total displacement $Z_i$.
\end{itemize}

Thus, for each simulated signal we have a vector $X_i^\star = (\pmb{\alpha}_i, \pmb{\lambda}_i, PGA_i,$ 
$V_i, D_i, E_i, L_i) \in \R^{13}$ of 13 real parameters. We want to know whether the maximum total displacement $Z_i$ of the structure is greater than a certain threshold, for example twice the elasticity limit $Y$.

\section{Binary classification}

\subsection{Preprocessing of the training data} \label{preprocess}

The signals whose maximum linear displacement is less than the elasticity limit $Y$ are not interesting, since we know they do not reach the threshold:
\[ L_i < Y \quad \Rightarrow \quad Z_i = L_i \quad \text{ and thus } \quad Z_i < Y. \]

This discarded $66\%$ of the simulated signals. We also discarded a few signals ($0.3\%$) whose maximum linear displacement was too high ($L_i > 6Y$), since the mechanical model we use is not realistic beyond that level. Therefore, we ended with a subset $I$ of our database such that:

\[ \forall i\in I \quad L_i \in [Y,6Y].\]

We have kept $N=33718$ simulated signals out of a total of $N_s = 10^5$. On those $N$ signals, a Box-Cox transform was applied on each component of $X_i^\star$. This non-linear step is critical for the accuracy of the classification, especially when we later use linear SVM classifiers. The Box-Cox transform is defined by:

\begin{equation} BC(x,\delta) = \left\lbrace \begin{array}{ll} \displaystyle \frac{x^\delta -1}{\delta} \quad & \text{ if } \delta \neq 0 \\
\log (x) \quad & \text{ if } \delta = 0. \end{array} \right. \end{equation}

\begin{figure}[h]
\centering
\subfloat[\label{histoa}]{\includegraphics[width=6cm]{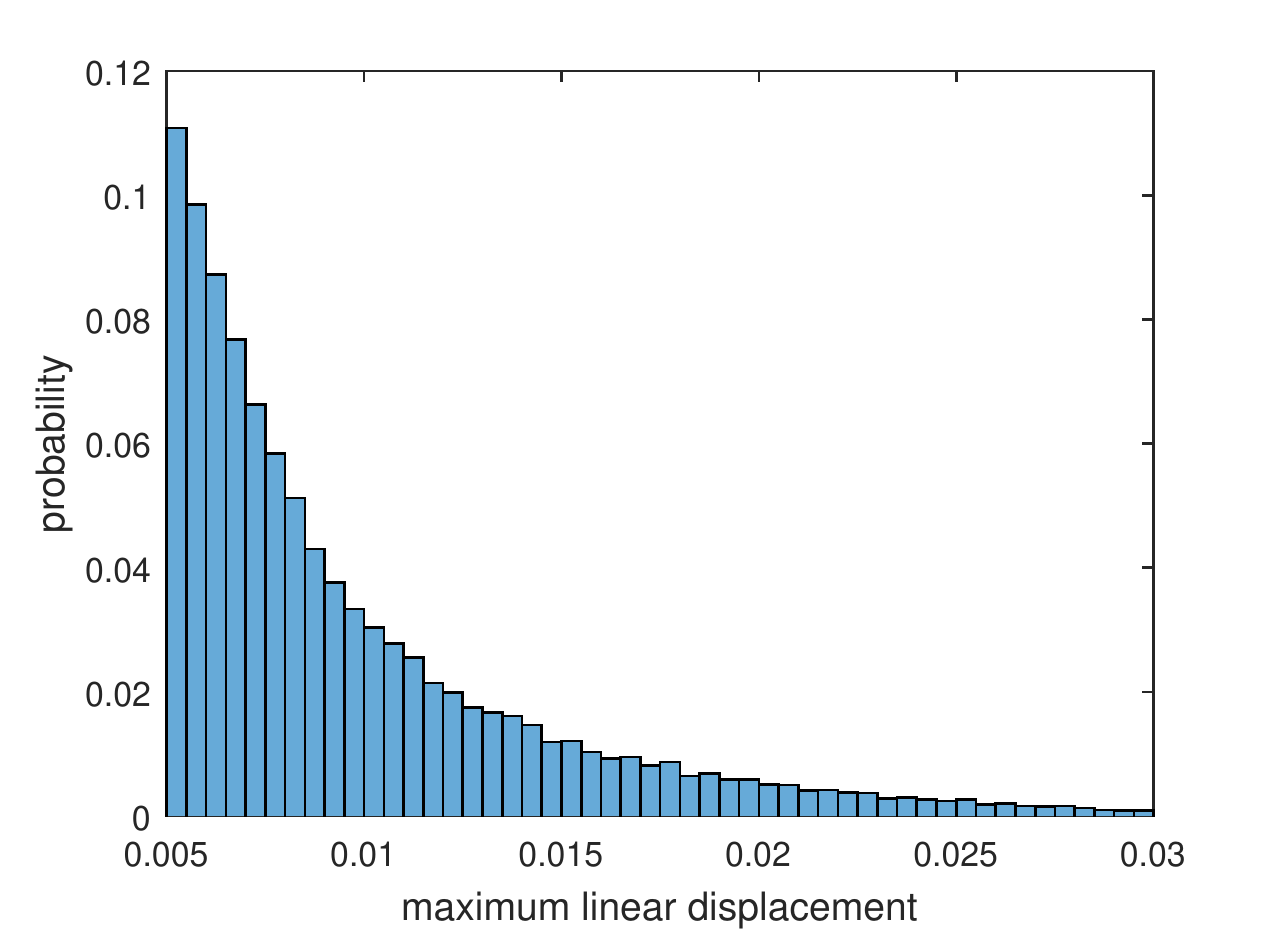}}
\subfloat[\label{histob}]{\includegraphics[width=6cm]{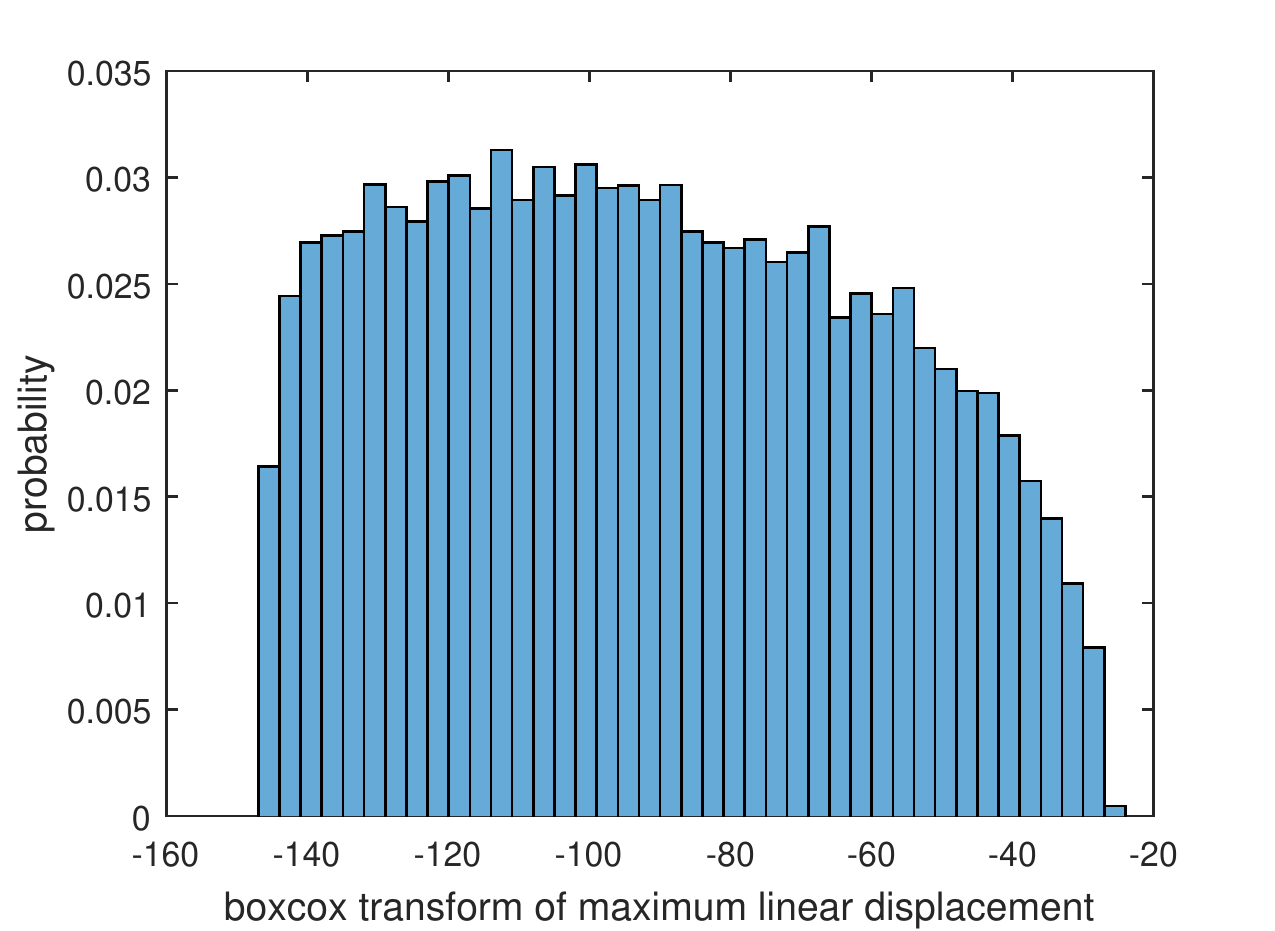}}
\caption{Histograms of $L$, (a) before and (b) after a Box-Cox transform with parameter $\delta = -0.928$.}
\label{histo}
\end{figure}

The parameter $\delta$ is optimized, for each component, assuming a normal distribution and maximizing the log-likelihood. Figure~\ref{histo} shows that $L$ is heavily modified by this transformation, with an optimal parameter of $\delta = -0.928$.

Finally, each of the 13 components was standardized, thus forming the training database $\mathcal{X} = \lbrace X_1,\dots,X_N \rbrace$ with $X_i \in \R^{13}$.

\subsection{Simple classifiers} \label{simple}

At the most basic level, a binary classifier is a labeling function

\begin{equation} \begin{array}{rcl} \hat{l}: \R^d & \longrightarrow & \lbrace -1;1\rbrace \\ X & \longmapsto & \hat{l}(X), \end{array} \end{equation}
that, given a vector $X \in \R^d$ corresponding to a seismic signal $s(t)$, gives us an estimated label $\hat{l}$. In our setting, the true label $l_i$ of instance $X_i$ is $1$ if the displacement $Z_i$ is greater than the damage threshold $2Y$, and $-1$ otherwise:

\begin{equation} l_i = \TR{\rm{sgn}(Z_i-2Y)} = \left\lbrace \begin{array}{rl} 1 \quad & \text{ if } \quad Z_i > 2Y,\\ -1 \quad & \text{otherwise}. \end{array} \right. \end{equation}

Note that the true label $l_i$ is not in general a function of the vector $X_i$, since it depends on the full signal $s_i(t)$ \TR{when} $X_i$ only gives us macroscopic measures of the signal; therefore, a perfect classifier $\hat{l}(X_i)$ may not exist.

\begin{figure}[h!]
\centering
\includegraphics[width=12cm]{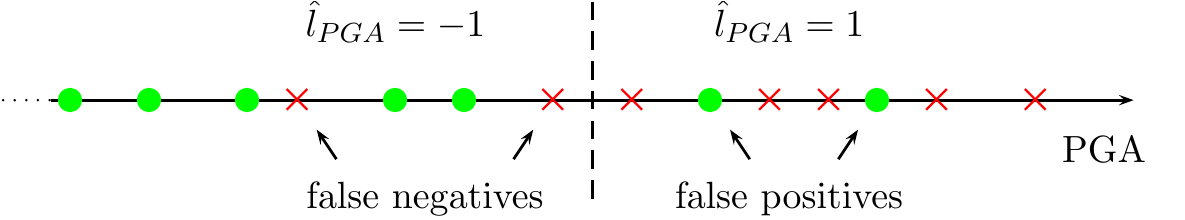}
%\begin{pspicture*}(-6,-1.2)(6,1)
%
%\psline[linestyle=dotted](-6,0)(-5.5,0)
%\psline{->}(-5.5,0)(5.5,0)
%\rput(5.2,-0.5){PGA}
%
%\psdots[dotstyle=*,dotsize=7pt,linecolor=green](-5.3,0)(-4.5,0)(-3.5,0)(-2,0)(-1.3,0)(1.2,0)(2.9,0)
%\psdots[dotstyle=x,dotsize=7pt,linecolor=red](4.5,0)(3.5,0)(-3,0)(-0.4,0)(0.4,0)(1.8,0)(2.4,0)
%
%\psline[linestyle=dashed](0,-1)(0,1)
%%\psline{->}(0.3,0.7)(1,0.7)
%\rput(2,0.8){$\hat{l}_{PGA} = 1$}
%%\psline{->}(-0.3,0.7)(-1,0.7)
%\rput(-2,0.8){$\hat{l}_{PGA} = -1$}
%
%\rput(-1.7,-1){false negatives}
%\psline{->}(-2.6,-0.6)(-2.8,-0.3)
%\psline{->}(-0.8,-0.6)(-0.6,-0.3)
%\rput(2,-1){false positives}
%\psline{->}(1.6,-0.6)(1.4,-0.3)
%\psline{->}(2.5,-0.6)(2.7,-0.3)
%
%\end{pspicture*}
\caption{Choice of the threshold for the binary PGA classifier $\hat{l}_{PGA}$.}
\label{M}
\end{figure}

One of the simplest choice for a classifier is to look at only one component of the vector $X$. For example, it is obvious that the PGA is highly correlated with the maximum total displacement $Z_i$; therefore, we can define the PGA classifier $\hat{l}_{PGA}$ as:

% \begin{equation} \hat{l}_{PGA}(X) = \left\lbrace \begin{array}{rl} 1 \quad & \text{if} \quad PGA > M, \\ -1 \quad & \text{otherwise}, \end{array} \right. 
% \end{equation}

\begin{equation} \hat{l}_{PGA}(X) =  \TR{\rm{sgn}(PGA - M)}
\end{equation}

where $M$ is a given threshold. Moving the threshold up results in less false positives ($\hat{l}=1$ when the real label is $l=-1$) but more false negatives ($\hat{l}=-1$ when the real label is $l=1$); and moving the threshold down results in the opposite. Therefore, there exists a choice of $M$ such that the number of false positives and false negatives are equal, as can be seen in figure \ref{M}.

Note that this choice does not guarantee that the \textit{total} number of misclassifications is minimal. Similarly, we can also define a classifier $\hat{l}_L$ based on the maximum linear displacement $L$, since the linear displacement is also highly correlated with the total displacement. These two simple classifiers give us a baseline to measure the performance of more advanced classifiers.

\subsection{SVMs and active learning}

\subsubsection{Support vector machines}

In machine learning, support vector machines (SVMs) are supervised learning models used for classification and regression analysis. In the linear binary classification setting, given a training data set $\lbrace X_1,\dots,X_n \rbrace$ that are vectors in $\R^d$, and their labels $\lbrace l_1,\dots,l_n \rbrace$ in $\lbrace -1,1 \rbrace$, the SVM is a hyperplane of $\R^d$ that separates the data by a maximal margin. More generally, SVMs allow one to project the original training data set $\lbrace X_1,\dots,X_n \rbrace$ onto a higher dimensional feature space via a Mercer kernel operator $K$. The classifier then associates to each new signal $X$ a score $f_n(X)$ given by:

\begin{equation} f_n(X) = \sum_{i=1}^n \alpha_i K(X_i, X). \end{equation}

A new seismic signal represented by the vector $X$ has an estimated label $\hat{l}$ of $1$ if $f_n(X)>0$, $-1$ otherwise. In a general SVM setting, most of the labeled instances $X_i$ have an associated coefficient $\alpha_i$ equal to $0$; the few vectors $X_i$ such that $\alpha_i \neq 0$ are called "support vectors", thus the name "support vector machine". This historical distinction among labeled instances is less relevant in the case of active learning (see next section), since most of the $\alpha_i$ are non-zero. In the linear case, $K(X_i,X)$ is just the scalar product in $\R^d$, and the score is:

\begin{equation} f_n(X) = W^T X + c, \label{eqW} \end{equation}
where $W \in \R^d$ and $c \in \R$ depend on the coefficients $\alpha_i$. Another commonly used kernel is the radial basis function kernel (or RBF kernel) $K(U,V) = e^{-\gamma (U-V) \cdot (U-V)}$, which induces boundaries by placing weighted Gaussians upon key training instances.

\subsubsection{Active learning}

Computing the total displacement $Z_i$ of the structure (and thus the label $l_i$) is very costly for a complex structure, limiting the size of the training data. Fortunately, it is possible to make accurate classifiers using only a limited number of labeled training instances, using active learning.

In the case of pool-based active learning, 
we have, in addition to the labeled set $\mathcal{L} = \lbrace X_1,\dots,X_n \rbrace$, access to a set of unlabeled samples $\mathcal{U} = \lbrace X_{n+1},\dots,X_N \rbrace$ (therefore we have $\mathcal{X} = \mathcal{L} \cup \mathcal{U}$). We assume that there exists a way to provide us with a label for any sample $X_i$ from this set (in our case, running a full simulation of the physical model using signal $s_i(t)$), but the labeling cost is high. After labeling a sample, we simply add it to our training set. In order to improve a classifier 
it seems intuitive to query labels for samples that cannot be easily classified. Various querying methods are possible \cite{Kre,Ton}, but the method we present here only requires to compute the score $f_n(X)$ for all samples in the unlabeled set, then to identify a sample that reaches the minimum of the absolute value $\vert f_n(X) \vert$, since a score close to $0$ means a high uncertainty for this sample. Thus, we start with  $n=2$ samples $j_1$ and $j_2$, labeled $+1$ and $-1$. Recursively, if we know the labels of signals $j_1,\dots,j_n$:
\begin{itemize}
\item we compute the SVM classifier associated with the labeled set $\lbrace (X_{j_1}, l_{j_1}),\dots,(X_{j_n}, l_{j_n}) \rbrace$;
\item for each unlabeled instance $X_i$, $i \in \llbracket 1,N \rrbracket \backslash \{j_1,\ldots, j_n\}$,
 we compute its score 
 \begin{equation} f_n(X_i)  = \displaystyle \sum_{k=1}^{n} \alpha_k K(X_{j_k}, X_i) ; \nonumber \end{equation}
\item we query the instance with maximum uncertainty for this classifier:
\begin{equation} j_{n+1} = \argmin_{i \in \llbracket 1,N \rrbracket \backslash \{j_1,\ldots, j_n\}}\vert f_n(X_i) \vert, \end{equation}
and compute the corresponding maximum total displacement $Z_{j_{n+1}}$ by running a full simulation of the physical model;
\item the instance $(X_{j_{n+1}}, l_{j_{n+1}}\TR{= \rm{sgn}(Z_{j_{n+1}}-2Y)})$ is added to the labeled set.\\
\end{itemize}

\subsubsection{Choice of the starting points}

The active learner needs two starting points, one on each side of the threshold. After the preprocessing step, about $17 \%$ of all remaining instances have a displacement greater than the threshold (although this precise value is usually unknown). It can be tempting to choose, for example, the signal with the smallest PGA as $j_1$ and the signal with the biggest PGA as $j_2$. However, running simulations with these signals is costly and give us a relatively useless information. We prefer to choose the starting points randomly, which also allows us to see how this randomness affects the final performance of the classifier.

The linear displacement $L_i$ and the $PGA_i$ of a signal are both obviously strongly correlated with the displacement $Z_i$. As a consequence, it is preferable that the starting points respect the order for these two variables:

\begin{equation} Z_{j_1} < 2Y < Z_{j_2}, \quad L_{j_1} < L_{j_2} \quad \text{and} \quad PGA_{j_1} < PGA_{j_2}. \label{start} \end{equation}

Indeed, if $j_1$ and $j_2$ are such that, for example, $Z_{j_1} < 2Y < Z_{j_2}$ but $PGA_{j_1} > PGA_{j_2}$, then the active learner starts by assuming that the PGA and displacement have a negative correlation, and it can take many iterations before it "flips"; in some rare instances the classifier performs extremely poorly for several hundreds of iterations. Thus, the starting points $j_1$ and $j_2$ are chosen such that equation (\ref{start}) is automatically true, using quantiles of the PGA and linear displacement. $j_1$ is chosen randomly among the instances whose PGA is smaller than the median PGA and whose linear displacement is smaller than the median linear displacement:

\begin{equation} j_1 \in \left\lbrace i \in \llbracket 1,N \rrbracket \quad \vert \quad PGA_i < D_5(PGA) \quad \& \quad L_i < D_5(L) \right\rbrace, \end{equation}

where $D_5(X)$ denotes the median of set $X$. It is almost certain that any instance in this set satisfies $Z_i < 2Y$ and thus $l_i = -1$. Similarly, $j_2$ is chosen using the 9th decile of both PGA and linear displacement:

\begin{equation} j_2 \in \left\lbrace i \in \llbracket 1,N \rrbracket \quad \vert \quad PGA_i > D_9(PGA) \quad \& \quad L_i > D_9(L) \right\rbrace, \end{equation}

where $D_9(X)$ denotes the 9th decile of a set $X$. The probability that $Z_i > 2Y$ in this case was found to be $97\%$. If we get unlucky and $Z_i < 2Y$ then we discard this signal and choose another one.

%We will now see how good our final classifier is, depending on the total number of labeled signals.

\subsection{ROC curve and precision/recall breakeven point}

The SVM classifier gives an estimated label $\hat{l}_i$ to each signal $s_i$ depending on its score $\hat{l}_i = \TR{\rm{sgn}(f_n(X_i))}$.
% \begin{equation} \hat{l}_i = \left\lbrace \begin{array}{ll} 1 \quad & \text{ if } f_n(X_i) \geq 0, \\ -1 \quad & \text{ if } f_n(X_i) < 0.\end{array} \right. \end{equation}
% \begin{equation} \hat{l}_i = \TR{\rm{sgn}(f_n(X_i))} \end{equation}
As for the simple classifiers (section \ref{simple}), we can set a non-zero limit $\beta$, and define the classifier as:

% \begin{equation} \hat{l}_i(\beta) = \left\lbrace \begin{array}{ll} 1 \quad & \text{ if } f_n(X_i) \geq \beta, \\ -1 \quad & \text{ if } f_n(X_i) < \beta,\end{array} \right. \quad \beta \in \mathbb{R}. \label{beta} \end{equation}

\begin{equation} \hat{l}_i(\beta) = \TR{\rm{sgn}(f_n(X_i)-\beta)} \label{beta}, \quad \beta \in \mathbb{R} \end{equation}

\begin{figure}[h]
\centering
\includegraphics[width=6.5cm]{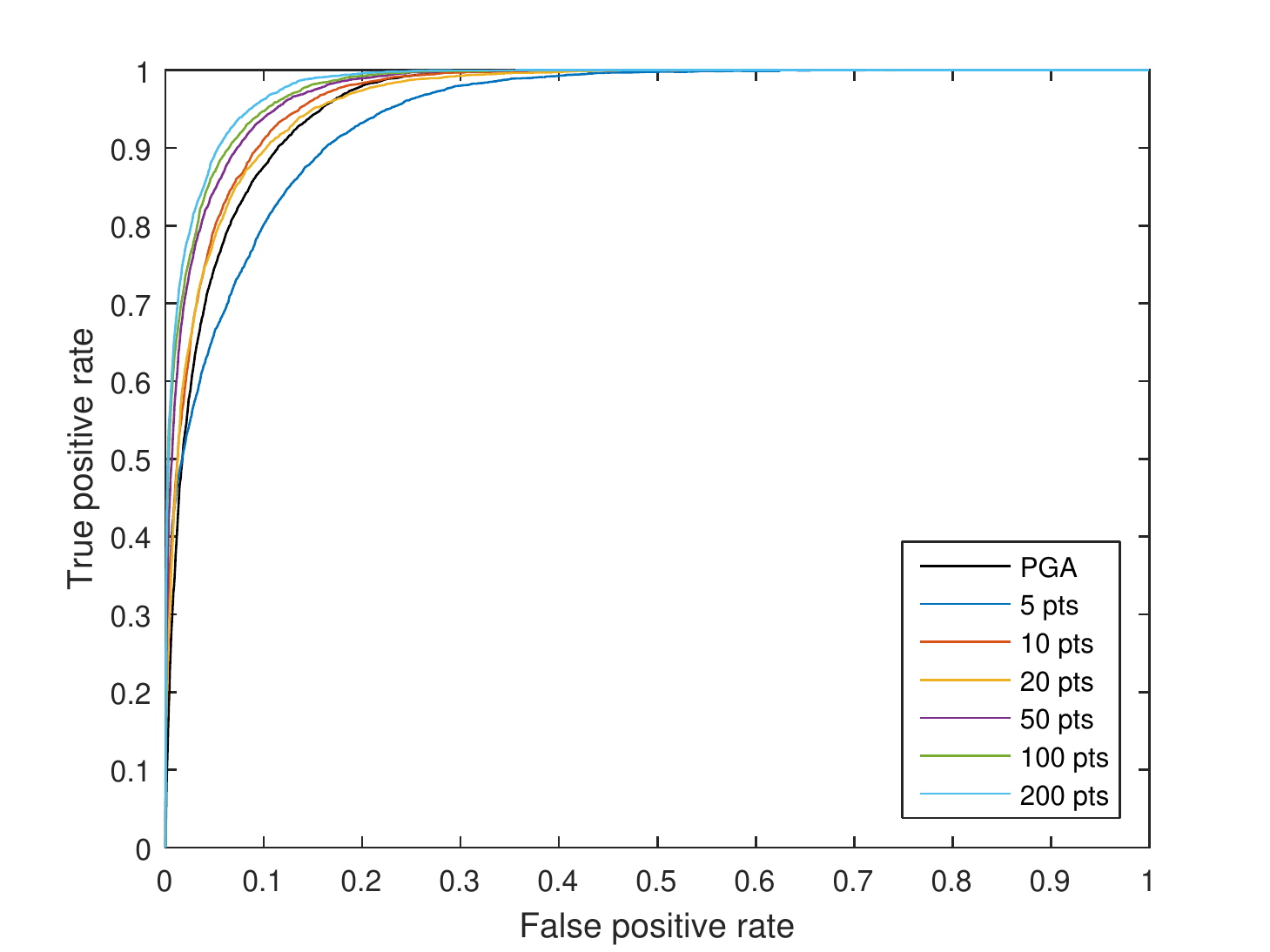}
\caption{ROC curves for the PGA classifier (black) and for 6 active learners after $n$ iterations ($n = 5,10,20,50,100$ and $200$).}
\label{exroc}
\end{figure}

If $\beta>0$, then the number of false positives ($l_i=-1$ and $\hat{l}_i=1$) is smaller, but the number of false negatives ($l_i=1$ and $\hat{l}_i=-1$) is bigger, relative to the $\beta=0$ case, and the opposite is true if we choose $\beta<0$. Taking all possible values for $\beta \in \R$ defines the \textit{receiver operating characteristic} curve, or ROC curve. The area under the ROC curve is a common measure for the quality of a binary classifier. The classifier is perfect if there exists a value of $\beta$ such that all estimated labels are equal to the true labels; in this case the area under the curve is equal to $1$. Figure \ref{exroc} shows one example of active learning, with ROC curves corresponding to different numbers of labeled signals. As expected, the classifier improves on average when the labeled set gets bigger; and the active learner becomes better than the simple PGA classifier as soon as $n \geq 10$.

 % (Joachims,1998)
Another metric can be used to measure performance: the precision / recall breakeven point \cite{Kre}. Precision is the percentage of samples a classifier labels as positive that are really positive. Recall is the percentage of positive samples that are labeled as positive by the classifier. By altering the decision threshold on the SVM we can trade precision for recall, until both are equal, therefore defining the precision/recall breakeven point. In this case the number of false positives and false negatives are equal (see figure \ref{M}). This value is very easy to obtain from a practical point of view. Let us denote by $N_+$ the number of instances where the displacement is greater than the threshold (on a total of $N$ signals in the database):

\begin{equation} N_+ = \# \left\lbrace i \in \llbracket 1,N \rrbracket \quad \vert \quad l_i = 1 \right\rbrace. \end{equation}

We sort all instances according to their score, i.e. we find a permutation $\sigma$ such that:

\begin{equation} f_n(X_{\sigma(1)}) \leq \dots \leq f_n(X_{\sigma(N)}). \end{equation}

Then the precision/recall breakeven point (PRBP) is equal to the proportion of positive instances among the $N_+$ instances with the highest score:

\begin{equation} \begin{array}{rll}  {\rm PRBP} & = & \displaystyle \frac{\# \left\lbrace i \in \llbracket 1,N \rrbracket \quad \vert \quad l_i = 1 \quad \& \quad \sigma(i) > N-N_+ \right\rbrace }{\# \left\lbrace i \in \llbracket 1,N \rrbracket \quad \vert \quad \sigma(i) > N-N_+ \right\rbrace} \\
& &\\
& = & \displaystyle \frac{\# \left\lbrace i \in \llbracket 1,N \rrbracket \quad \vert \quad l_i = 1 \quad \& \quad \sigma(i) > N-N_+ \right\rbrace }{N_+}  \end{array} \label{prbp}  \end{equation}

This criteria does not depend on the number of true negatives (unlike the false positive rate, used in the ROC curve). In particular, it is not affected by our choice of preprocessing of the training data, where we discarded all the weak signals ($L_i<Y$). (both metrics are affected by our choice to discard the very strong signals ($L_i>6Y$), but the effect is negligible in both cases).

\subsection{Results}

We now compare different classifiers with the precision/recall breakeven point. More precisely, we compare different \textit{orderings} of all signals, since only the order matters to the PRBP; for instance, the PGA does not give directly a label, but we can compute the PRBP of the PGA classifier with equation \ref{prbp} using the permutation $\sigma_{PGA}$ that sorts the PGA of all signals. We can thus compare:
\begin{enumerate}
\item the simple PGA and maximum linear displacement classifiers $\hat{l}_{PGA}$ and $\hat{l}_L$, defined in section \ref{simple};
\item neural networks, trained with all instances and all labels (ie, with the $N=33718$ signals and labels), with either all 13 parameters, or just 4 of them: $(L, PGA, V, \omega_0)$ (the linear displacement, peak ground acceleration, peak ground velocity and filter frequency, see section \ref{reduction} for justification of this choice);
\item SVMs given by our active learning methods.
\end{enumerate}

\begin{figure}[!ht]
\centering
\subfloat[\label{res}]{\includegraphics[width=9.1cm]{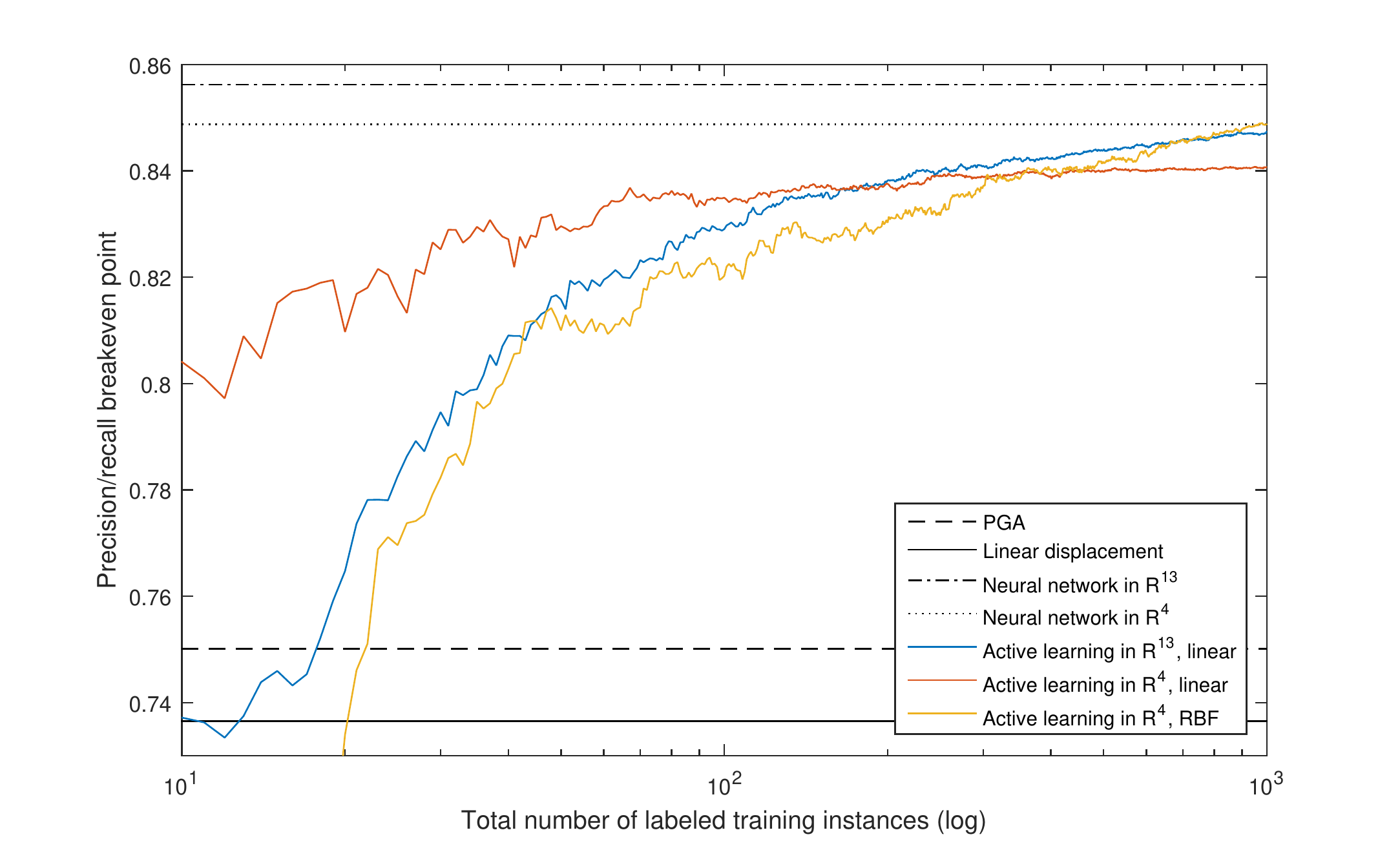}}\\
\subfloat[\label{variability}]{\includegraphics[width=9.5cm]{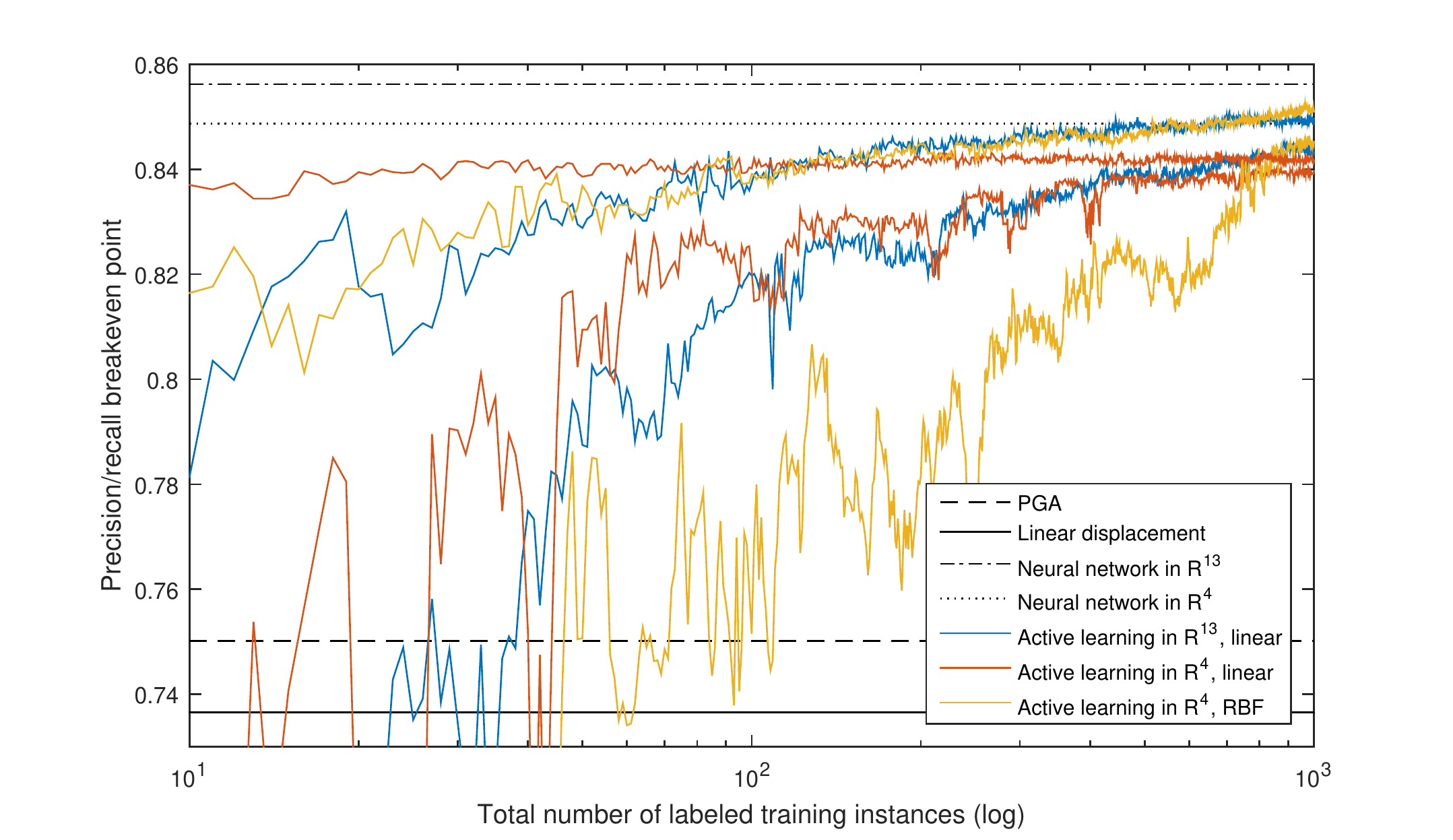}}
\caption{Performances of 3 active learning classifiers over 20 test cases. (a) Average. (b) Worst and best.}
\label{rest}
\end{figure}

\TR{The neural networks we used are full-connected Multi Layered Perceptrons (MLPs) with 2 layers of 26 and 40 neurons for $X \in \R^4$, and two layers of 50 and 64 neurons for $X \in \R^{13}$. In the active learning category, the performance depends on the number of iterations (between 10 and 1000). So, the results shown in figure \ref{rest} are functions of the number $n$ of labeled training instances, in logarithmic scale. The simple classifiers and the neural networks are represented as horizontal lines, since they do not depend on $n$. Moreover, this figure shows the PRBP of (i) a linear SVM using all 13 parameters (in blue), (ii) a linear SVM using only 4 parameters $(L, PGA, V, \omega_0)$ (in red) and (iii) a radial basis function (RBF) SVM, using the same 4 parameters (in yellow). As active learners depend on the choice of the first two samples, results of figure \ref{res} are obtained choosing 20 pairs of starting points $(j_1, j_2)$, then averaging the performance, knowing that the same starting points were used for all three types of SVMs. For completeness, figure~\ref{variability} shows the worst and best performances of the 3 classifiers on the 20 test cases.}

\newpage

Figure \ref{res} shows that active learning gives a much better classifier than the standard practice of using a single parameter (usually the PGA). The linear SVM with only 4 variables has initially the best performance on average, up to 150/200 iterations. The full linear SVM with 13 variables is better when the number of iterations is at least 200. The RBF kernel in $\R^4$ appears to have the best performance with 1000 labeled instances, outperforming the neural network in $\R^4$; however, it has a higher variability, as can be seen in figure \ref{variability}. Radial basis function SVMs with 13 parameters seem to always perform very poorly, and are not represented here. Figure \ref{variability} shows the lowest and highest score of all 20 test cases, independently for each number of iterations (one active learner can perform poorly at some point, and much better later, or the other way around). So, in conclusion, (i) active learners need a minimum of 30-40 iterations, otherwise they can end up worse than using the simpler PGA classifier, (ii) between 50 and 200 iterations, the linear SVM in $\R^4$ is the best choice, and has a relatively small variability and (iii) the RBF kernel seems quite unpredictable for less than 1000 iterations, and its performance depends wildly on the starting points, probably because of over-fitting.

\subsection{Results for different settings}

\begin{figure}[h]
\centering
\subfloat[\label{res10}]{\includegraphics[width=6.5cm]{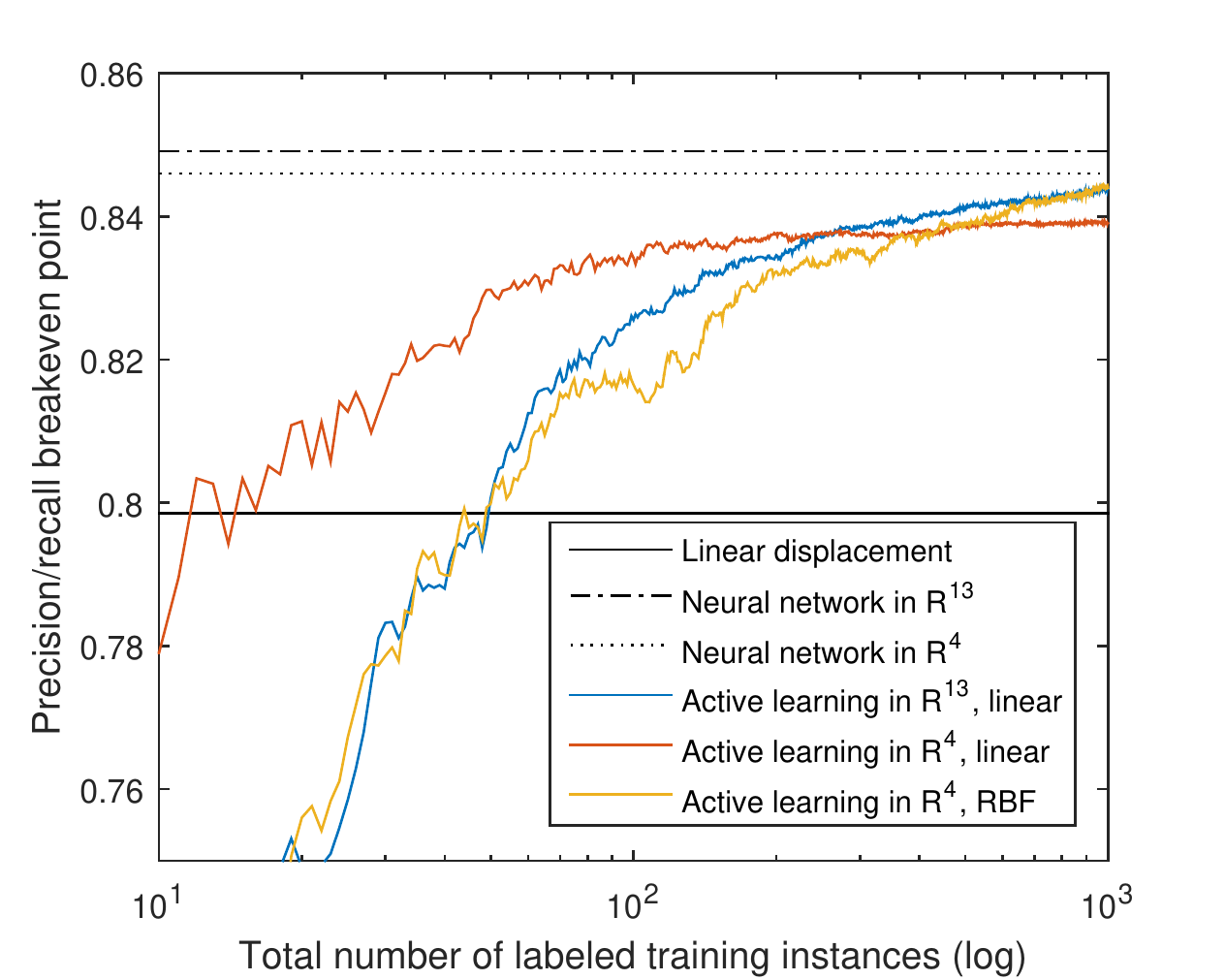}}
\subfloat[\label{res2.5}]{\includegraphics[width=6.5cm]{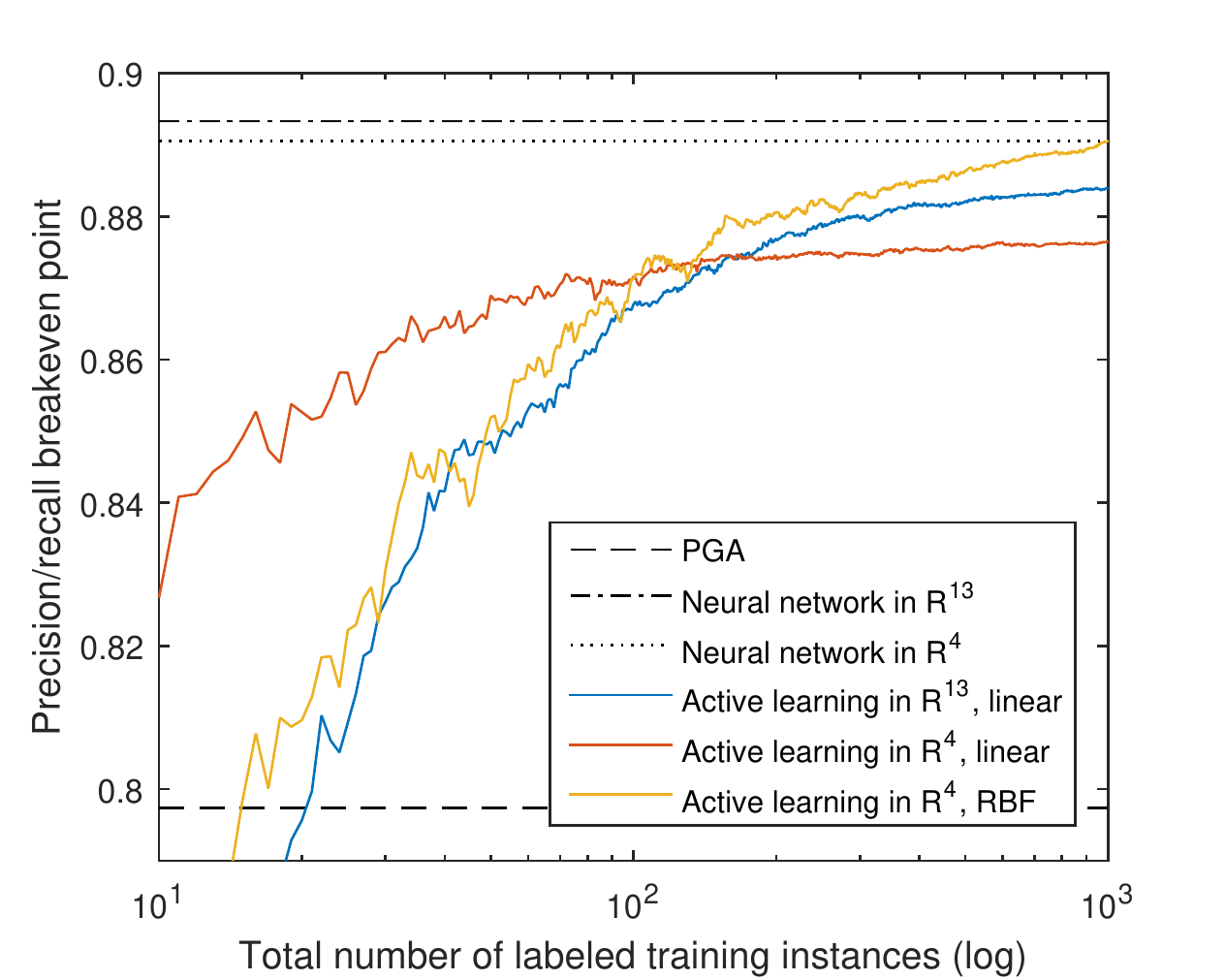}}
\caption{Average performance of active learning classifiers at (a) 2.5 Hz and (b) 10 Hz, over 20 test cases.}
\label{res10/2.5}
\end{figure}

Our methodology is very general and can be applied to a variety of structures. As an example, we compared the same classifiers on two structures with two different main frequencies, $2.5$ Hz and $10$ Hz, instead of the original $5$ Hz. The elasticity limit $Y$ was also changed so that approximately one third of all signals result in inelastic displacement: $Y=9\cdot 10^{-3}$ m for the $2.5$ Hz structure, $Y=5\cdot 10^{-3}$ m for $5$ Hz and $Y=1\cdot 10^{-3}$ m for $10$ Hz. The failure threshold was always chosen as $2Y$, which resulted in about $8.8\%$ of all signals attaining it for the $2.5$ and $10$ Hz cases, compared to $5.7\%$ in the $5$ Hz setting. As shown in figure~\ref{res10/2.5}, the performances of active learners are very similar to the 5 Hz case, and the same conclusions apply. The performances of classifiers based on a single parameter, on the other hand, can vary a lot depending on the frequence of the structure: the PGA classifier provides a good classifier at high frequency (PRBP$=0.797$ at 10 Hz) but performs poorly at low frequency ($0.6$ at 2.5 Hz, it does not appear in figure \ref{res10/2.5}); while the linear displacement does the opposite (PRBP$=0.798$ at $2.5$ Hz, but PRBP$=0.69$ at 10 Hz). These results show that the active learning methodology is not just more precise, but also more flexible than the simple classifiers, and that with just 50 to 200 simulations it approaches the performance of a neural network using $33718$ simulations.

\subsection{Remark about the dimension reduction} \label{reduction}

\begin{figure}[!ht]
\centering
\includegraphics[width=9cm]{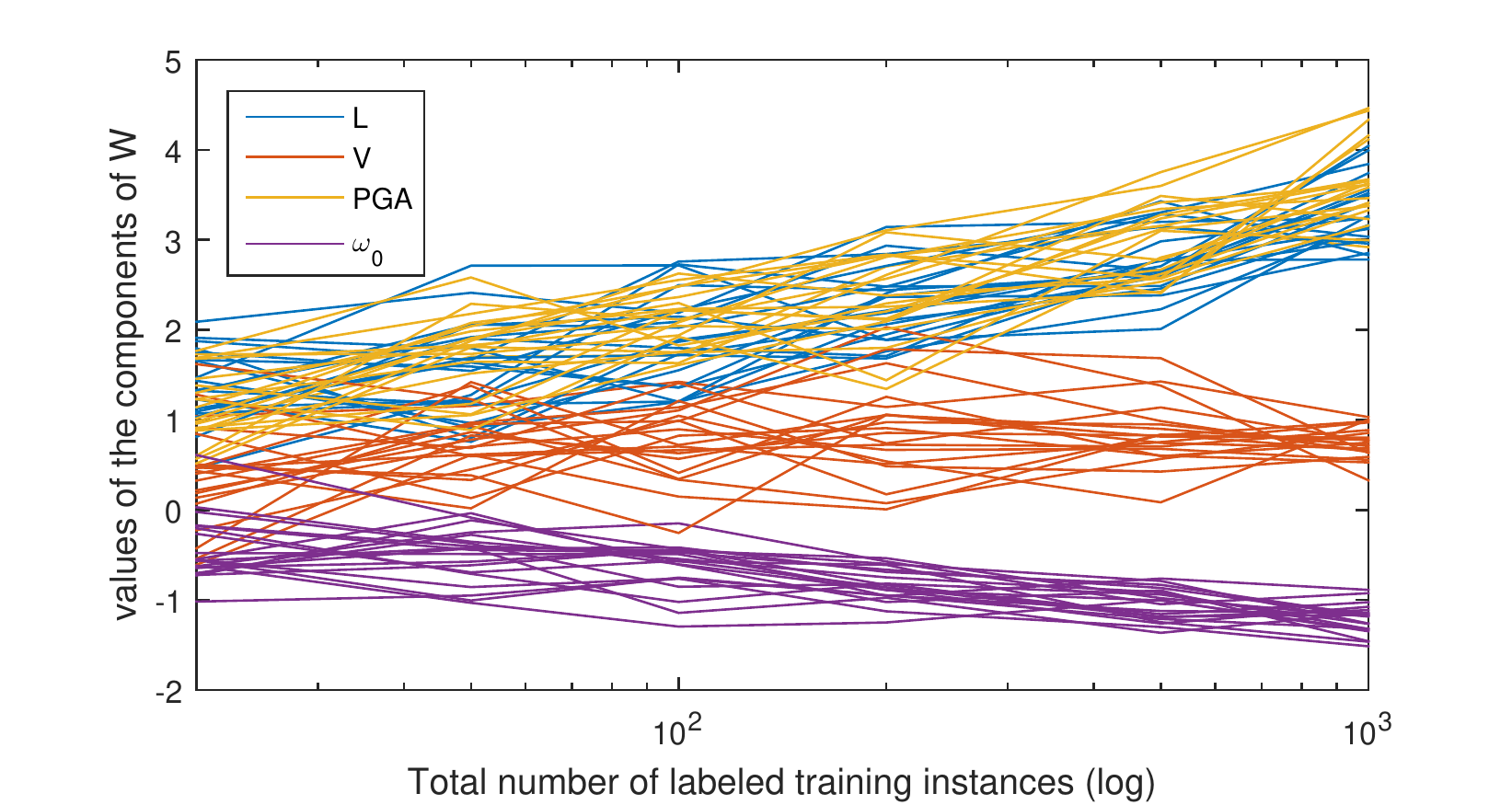}
\caption{Evolution of the 4 main components of $W$ for all 20 test cases.}
\label{W}
\end{figure}

In the linear case the score is equal to the distance to a hyperplane: $f_n(X) = W^T X + c$ (see equation \ref{eqW}). Therefore, we can see which of the components of $X$ are the most important for the classification simply by looking at the values of the components of $W$. Figure \ref{W} shows that the values of $W$ are roughly the same for all 20 test cases. After 1000 iterations, the coefficients for the PGA and maximum linear displacement $L$ end up between $3$ and $4$, the value for the maximum velocity $V$ is around $1$, and the value for the signal main frequency $\omega_0$ is around $-1$. The other 9 components of $W$ (when working with $X \in \R^{13}$) are all between $-1$ and $1$, but are smaller (in absolute value) than these 4 components. Note that comparing the values of the components of $W$ is only possible because the components of $X$ are standardized in the first place (cf. the preprocessing in section~\ref{preprocess}). As can be seen in the previous sections, reducing the dimension from $13$ to $4$ allows for a faster convergence, although the converged classifier is usually less precise. Continuing the active learning after 1000 iterations changes only marginally the results; even with $X \in \R^{13}$, both the PRBP and the values of $W$ stay roughly the same between 1000 and 5000 iterations.

\newpage

\section{Fragility curves}

SVM classifiers give to each signal $i$ a score $f_n(X_i)$ whose sign expresses the estimated label, but it does not give directly a probability for this signal to be in one class or the other. In order to define a \textit{fragility curve}, that is, the probability of exceeding the damage threshold as a function of a parameter representing the ground motion, we first need to assign a probability to each signal.

\subsection{Score-based fragility curve}

This probability depends only on the score $f_n(X)$. For a perfect classifier, the probability would be $0$ if $f_n(X)<0$ and $1$ if $f_n(X)>0$; for our SVM classifiers we use a logistic function:

\begin{equation} p(X) = \frac{1}{1+e^{-af_n(X)+b}}, \label{p(x)}\end{equation}
where $a$ and $b$ are the slope and intercept parameters of the logistic function ($b$ should be close to $0$ if the classifier has no bias, giving a probability of $1/2$ to signals with $f_n(X) \approx 0$). These parameters are computed using a logistic regression on the labeled set $\lbrace (X_{j_1}, l_{j_1}),\dots,(X_{j_n}, l_{j_n}) \rbrace$. To compare this estimation with the empirical failure probability of signals with a given score, we divide our database $\mathcal{X}$ in $K$ groups $(I_1,...,I_K)$ depending on their score, with the k-means algorithm; then we define the estimated and reference probabilities of each group:

\begin{equation} \begin{array}{rcl} p_k^{est} &=& \frac{1}{n_k} \sum_{i \in I_k} p(X_i),\\p_k^{ref} &=& \frac{1}{n_k}\# \left\lbrace i \in I_k \vert l_i=1 \right\rbrace , \end{array} \quad \text{ with } \quad n_k = \# I_k.\end{equation}

\begin{figure}[!ht]
\centering
\includegraphics[width=8.5cm]{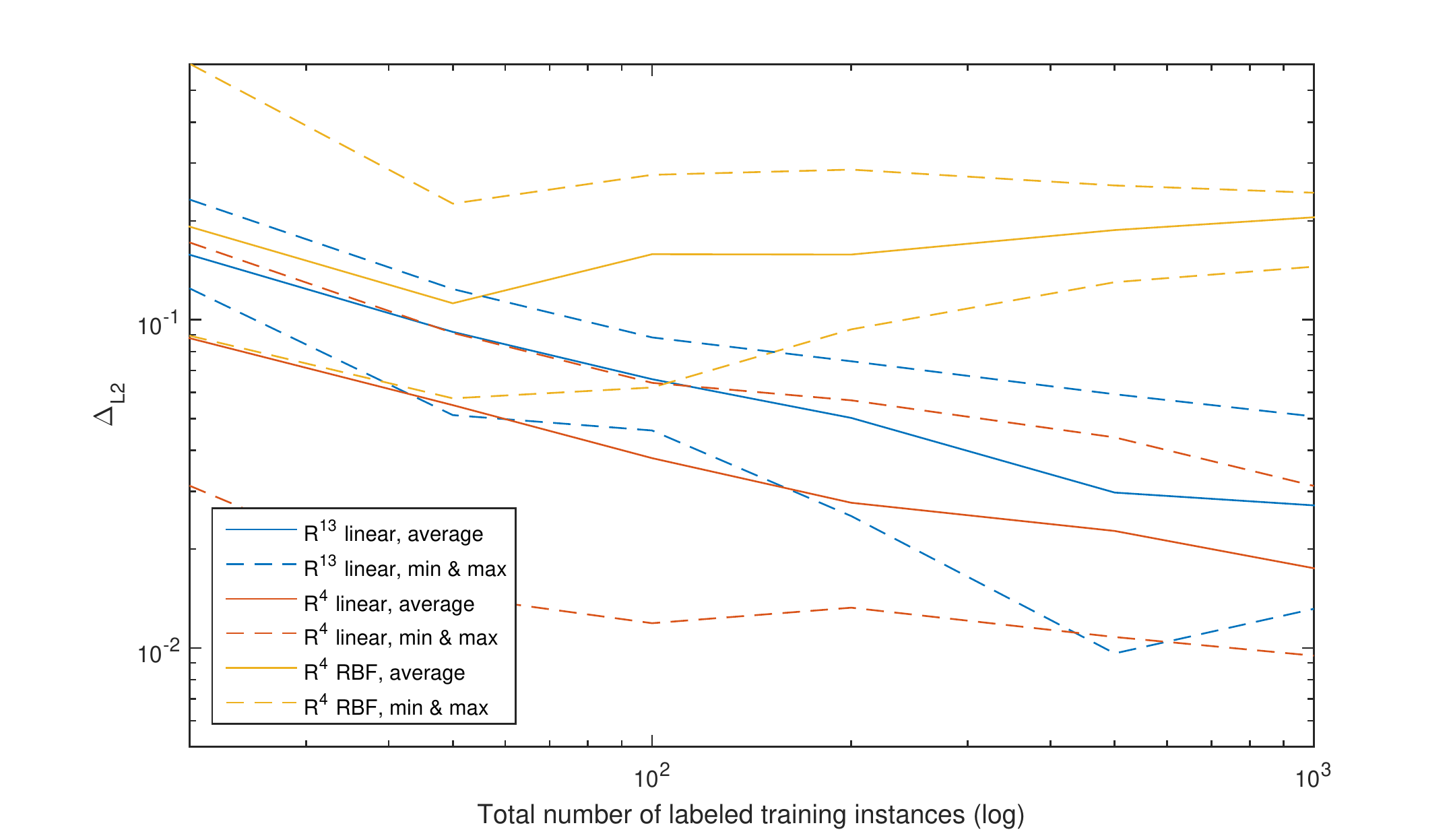}
\caption{Distance between the reference and estimated fragility curves for 3 different active learners.}
\label{precision}
\end{figure}

We can now compute the discrete $L_2$ distance between these two probabilities:

\begin{equation} \Delta_{L_2} = \sqrt{\frac{1}{N} \sum_{k=1}^K n_k (p_k^{ref} - p_k^{est})^2 }, \end{equation}
with $N = \sum_{k=1}^K n_k$. Figure \ref{precision} shows this distance for different classifiers using 20, 50, 100, 200, 500 and 1000 labeled instances. The three classifiers (linear SVM in $\R^{13}$, linear SVM in $\R^4$, and RBF kernels in $\R^4$) are compared on 20 test cases, using 20 pairs of starting points (the same for all three). The solid lines show the average $L_2$ errors, and the dashed lines show the minimum and maximum errors among all test cases.

\newpage

The average error goes down from $15\%$ after 20 iterations to less than $3\%$ after 1000 iterations for the linear SVM in $\R^{13}$, and from $9\%$ to less than $2\%$ for the linear SVM in $\R^4$. For the SVM using RBF kernels (in yellow in figure \ref{precision}), the average error does not decrease as the number of iterations increases, and ends up around $20\%$ after 1000 iterations. Figure \ref{4rbf} shows typical examples of fragility curves obtained with each method after 100 and 1000 iterations. Recall that the logistic functions (in red) are not fitted using all the real data (in blue), but only the labeled set, ie 100 or 1000 signals. The linear SVM in $\R^4$ has the least errors in terms of probabilities, although its PRBP is smaller than the linear SVM in $\R^{13}$ when using 1000 labeled instances.\\

\begin{figure}[!ht]
\centering
%\includegraphics[scale=0.57]{fragility13lin100z.pdf}
%\includegraphics[scale=0.57]{fragility13lin1000z.pdf}
%\caption{Real and estimated fragility curves using a linear SVM in $\R^{13}$, with $n=100$ (left) and $n=1000$ (right) labeled instances.}
%\label{13lin}
%\end{figure}
%
%\begin{figure}[h]
%\centering
%\includegraphics[scale=0.57]{fragility4lin100z.pdf}
%\includegraphics[scale=0.57]{fragility4lin1000z.pdf}
%\caption{Real and estimated fragility curves using a linear SVM in $\R^4$, with $n=100$ (left) and $n=1000$ (right) labeled instances.}
%\label{4lin}
%\end{figure}
%
%\begin{figure}[h]
%\centering
\subfloat[\label{4rbfa}]{\includegraphics[scale=0.56]{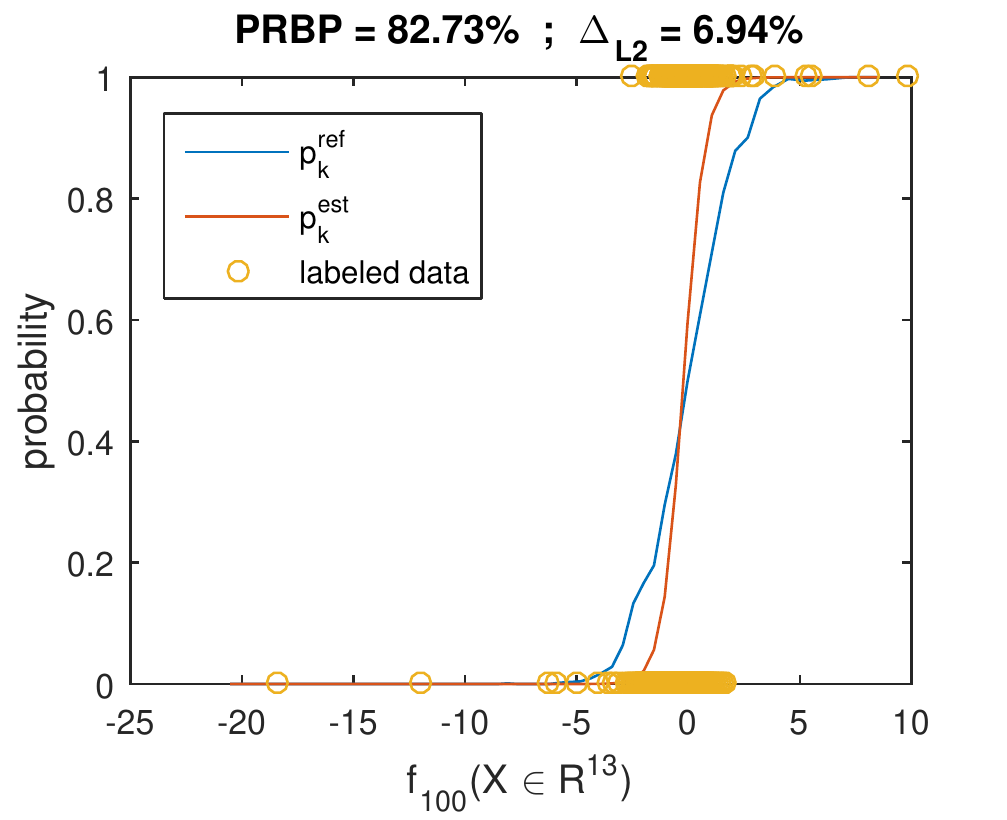}}
\subfloat[\label{4rbfb}]{\includegraphics[scale=0.56]{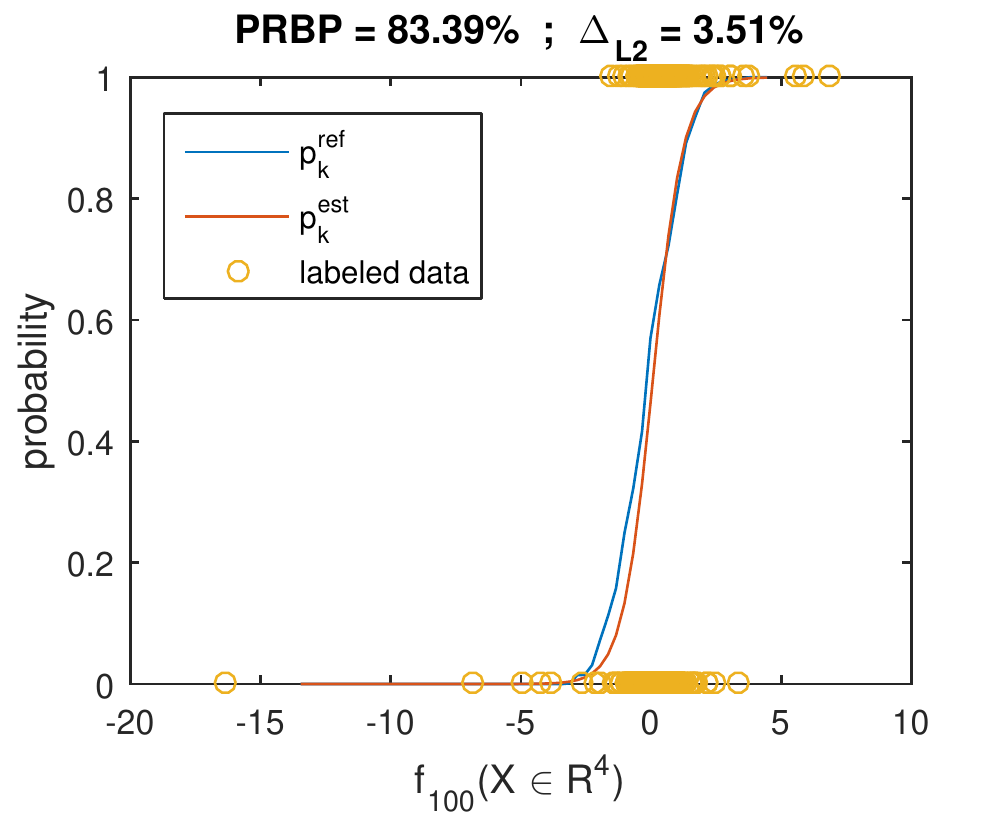}}
\end{figure}

\addtocounter{figure}{+1}
\begin{figure}[!ht]\ContinuedFloat
\centering
\subfloat[\label{4rbfc}]{\includegraphics[scale=0.56]{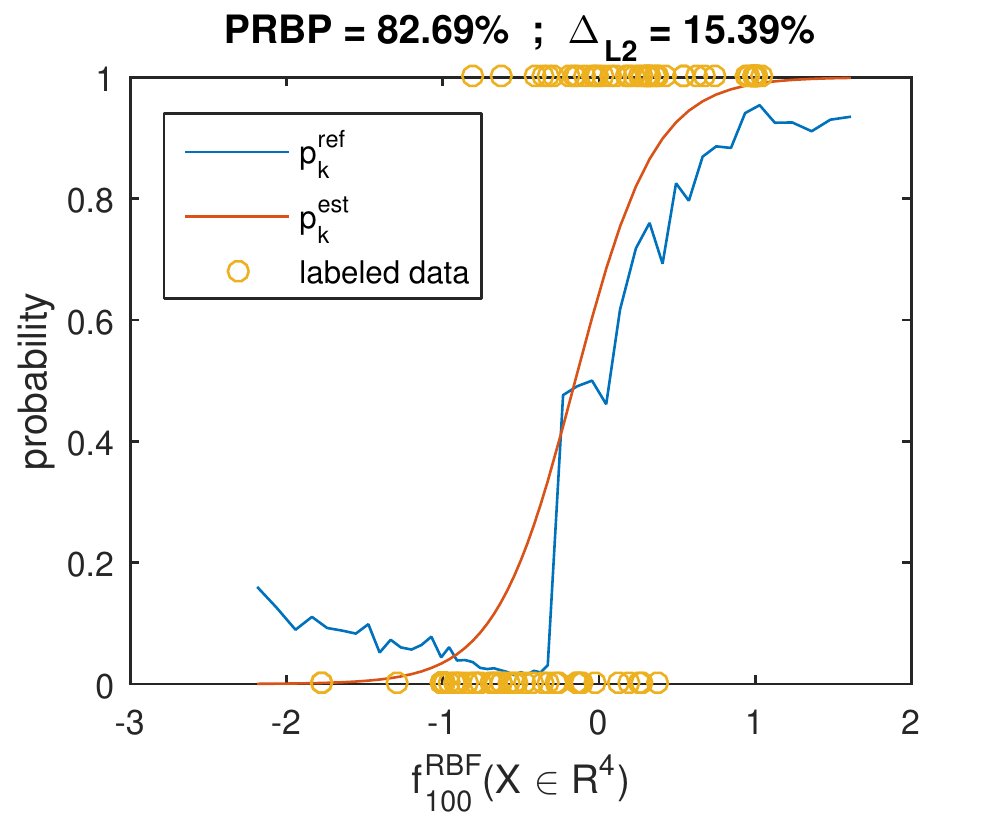}}
\subfloat[\label{4rbfd}]{\includegraphics[scale=0.56]{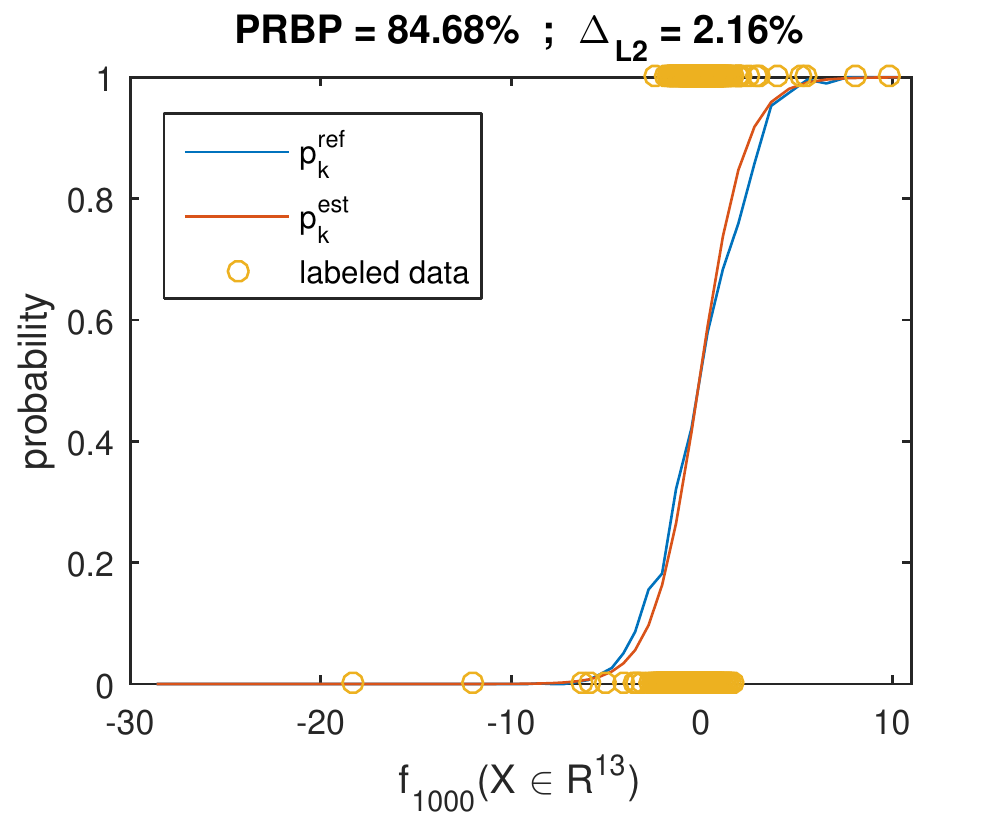}}\\
\subfloat[\label{4rbfe}]{\includegraphics[scale=0.56]{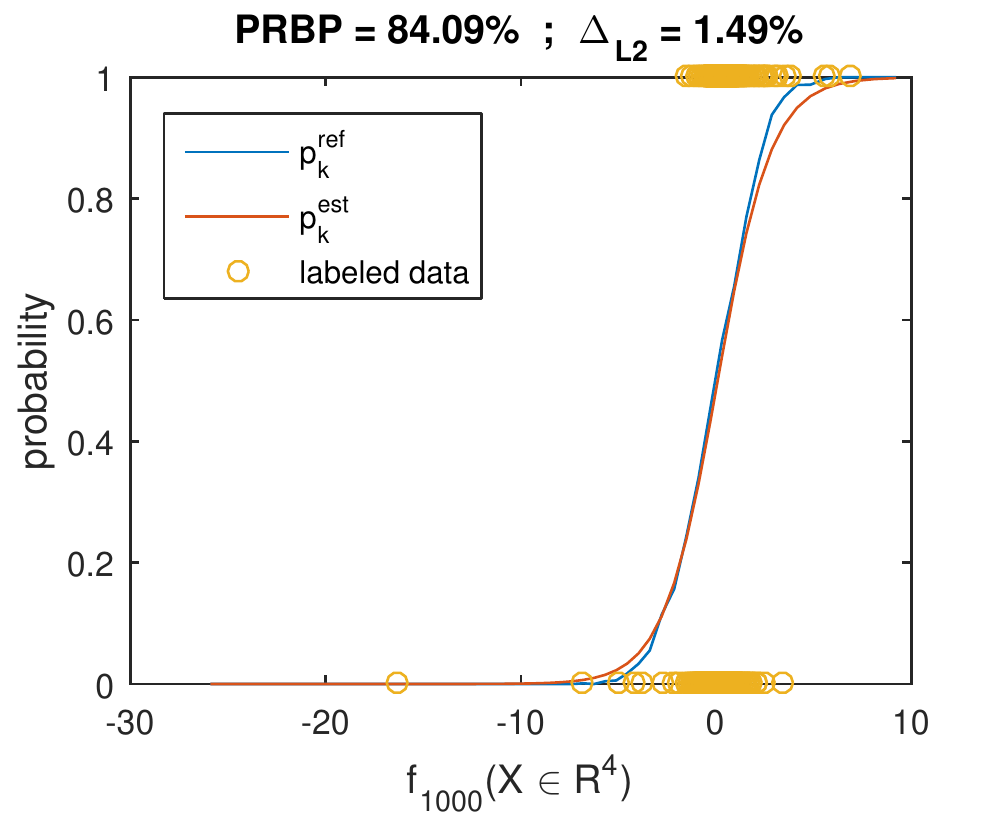}}
\subfloat[\label{4rbff}]{\includegraphics[scale=0.56]{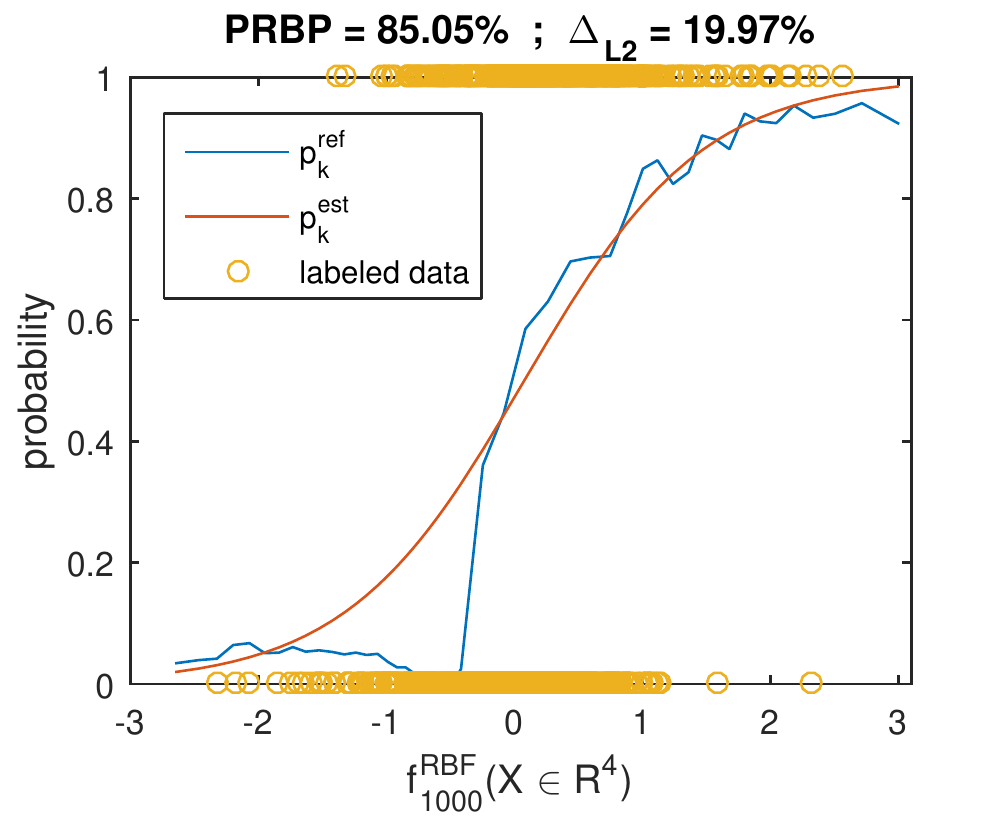}}
\caption{Reference and estimated fragility curves using  (a and d) a linear SVM in $\R^{13}$, (b and e) a linear SVM in $\R^4$ and (c and f) a RBF SVM in $\R^4$, with (a, b and c) $n=100$ or (d, e and f) $n=1000$ labeled instances.}
\label{4rbf}
\end{figure}

\newpage

\begin{figure}[!h]
\centering
\subfloat[\label{z(f)a}]{\includegraphics[width=6.5cm]{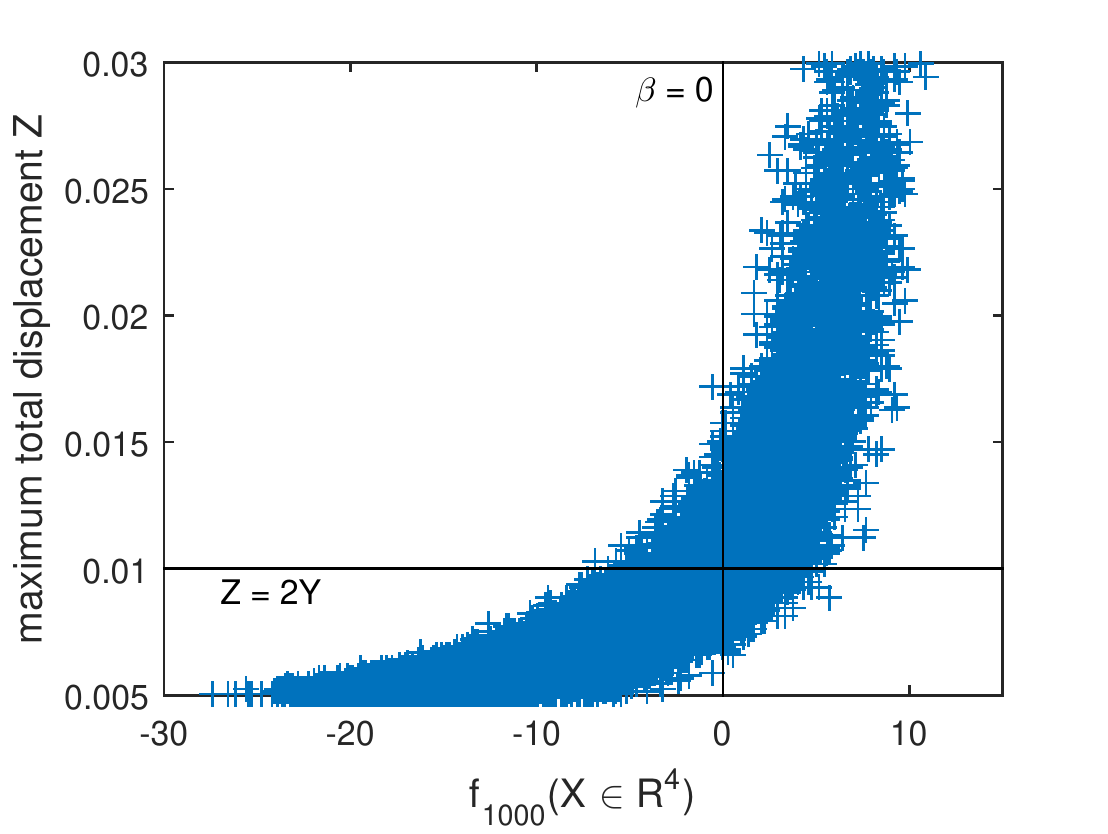}}
\subfloat[\label{z(f)b}]{\includegraphics[width=6.5cm]{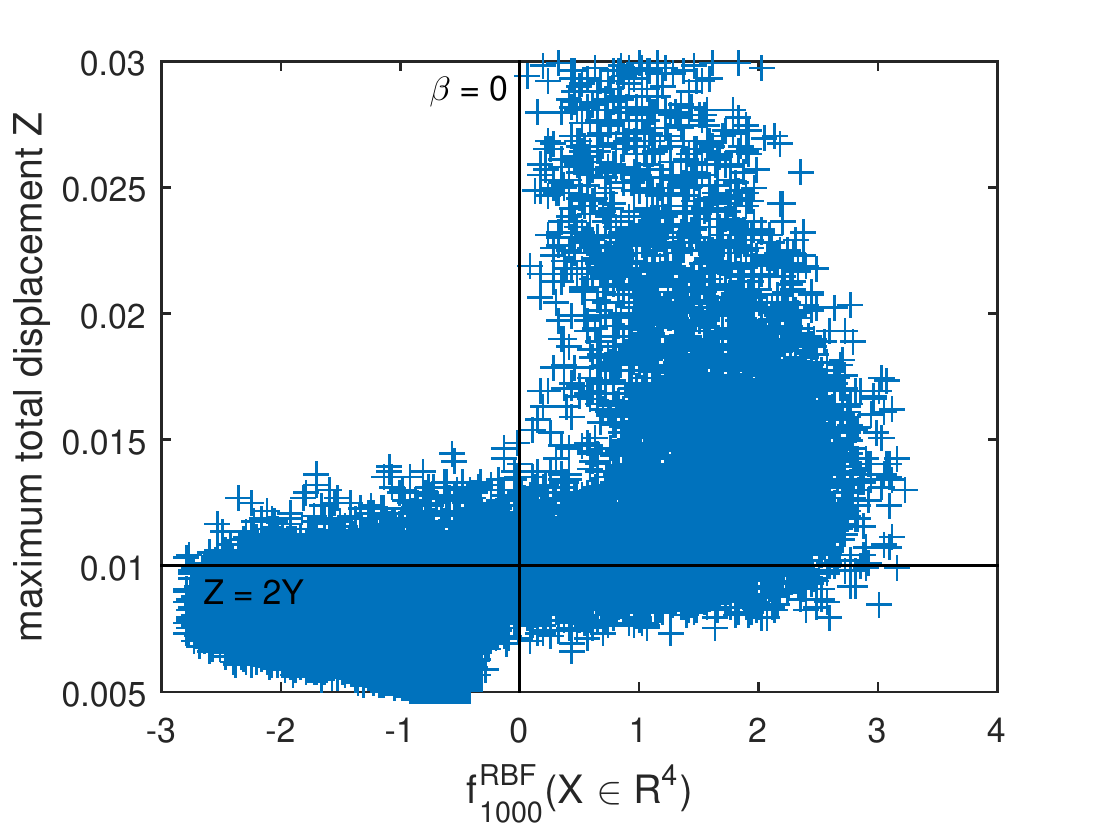}}
\caption{$Z$ as a function of the score given after 1000 iterations by (a) the linear SVM in $\R^4$ and (b) the RBF SVM in $\R^4$.}
\label{z(f)}
\end{figure}

The radial basis function kernel shows a very "strange" behaviour. The probability of failure is not even an increasing function of the score (figures \ref{4rbfc} and \ref{4rbff}); in particular, signals with a very negative score still have a $5-10\%$ chance of exceeding the threshold. This strange shape of $p_k^{ref}$ explains why the $\Delta_{L_2}$ error of RBF kernels is so high (figure \ref{precision}), since we tried to fit a logistic curve on a non-monotonous function. The reason for this major difference between linear and RBF kernels can be understood if we look at the maximum total displacement $Z$ as a function of the score $f_n(X)$, using both kernels (see figure \ref{z(f)}). Let us keep in mind that the RBF classifier at 1000 iterations is the most precise of all our active learners; it has the fewest false positives and false negatives of all (see table \ref{confusionmatrix}). The sign of the RBF score is thus an excellent predictor for binary classification.

\begin{table}[!ht]
\centering
  \begin{tabular}{| c | c | c |}
   \hline
   linear kernel & $f(X)<0$ & $f(X)>0$ \\ \hline
   $Z > 2Y$ & 1020 & 4711 \\ \hline
   $Z < 2Y$ & 27175 & 812 \\ \hline
  \end{tabular}  
  \quad
%	\vspace{0.5cm}    
  \begin{tabular}{| c | c | c |}
   \hline
   RBF kernel & $f(X)<0$ & $f(X)>0$ \\ \hline
   $Z > 2Y$ & 1009 & 4722 \\ \hline
   $Z < 2Y$ & 27287 & 700 \\ \hline
  \end{tabular}
\caption{Confusion matrix after 1000 iterations for the linear SVM in $\R^4$ (left) and the RBF SVM in $\R^4$ (right).}\label{confusionmatrix}
\end{table}

%However, when we use the score to predict a probability (eq \ref{p(x)})
Figure \ref{z(f)} shows that for the linear classifier, the score is a good predictor of the maximum total displacement $Z$, with a monotonous relation between the two; therefore the probability that a $Z>2Y$ is well-approximated by a logistic function of the score. The RBF score, on the other hand, is a poor predictor of the probability of failure, since the relation between the score and the maximum total displacement $Z$ is not monotonous. We can now understand the very high $\Delta_{L_2}$ errors for RBF kernels. Looking at figure \ref{z(f)}, we can see that the weakest signals ($Z = 0.005$, just above the elasticity limit) have a RBF score between $-1$ and $-0.4$. Since these weak signals are very common in our database, the reference probability $p_k^{ref}$ goes rapidly from $0.5$ for $f_n^{\text {RBF}}(X)=0$ to almost $0$ for $f_n^{\text {RBF}}(X) = -0.5$ (see figures \ref{4rbfc} and \ref{4rbff}), not because the number of positive signals changes significantly between $f_n^{\text {RBF}}(X)=-0.5$ and $f_n^{\text {RBF}}(X)=0$, but because the number of negative signals is more than 20 times bigger. The linear kernels do not have this problem, and therefore have much lower $\Delta_{L_2}$ errors. 

\begin{figure}[!ht]
\centering
\subfloat[\label{z(f)a}]{\includegraphics[width=6.5cm]{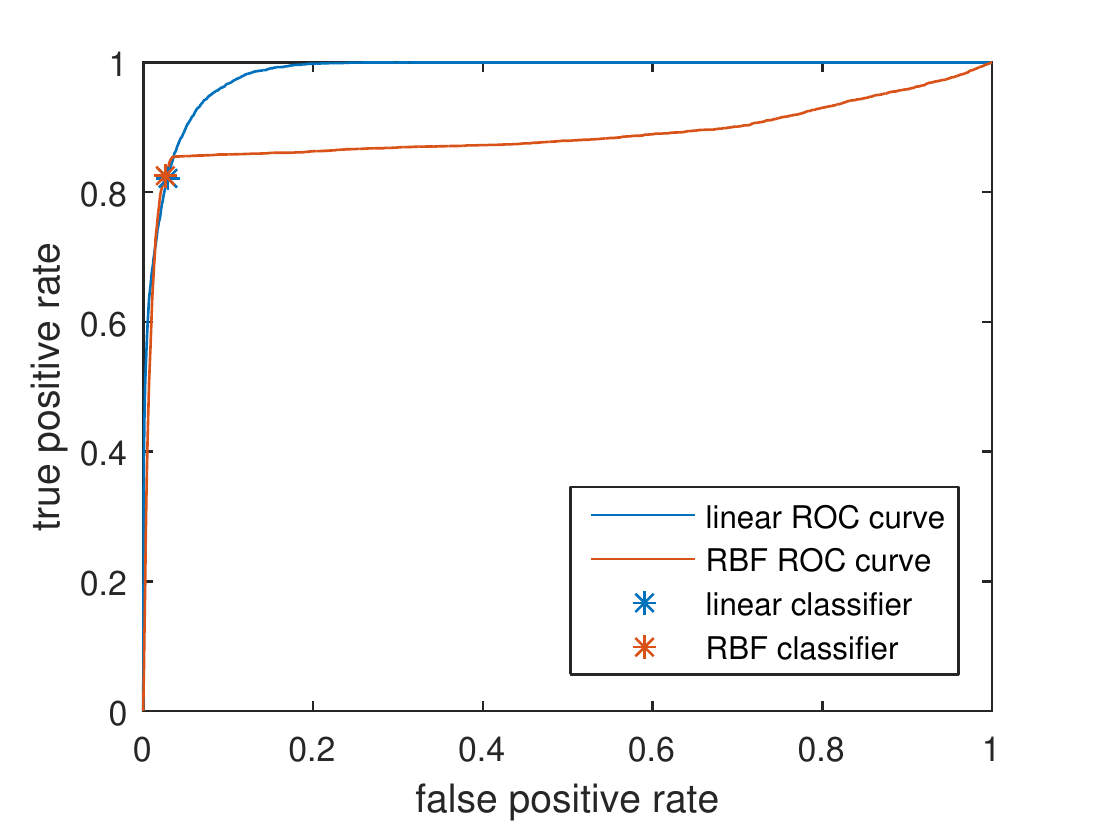}}
\subfloat[\label{z(f)a}]{\includegraphics[width=6.5cm]{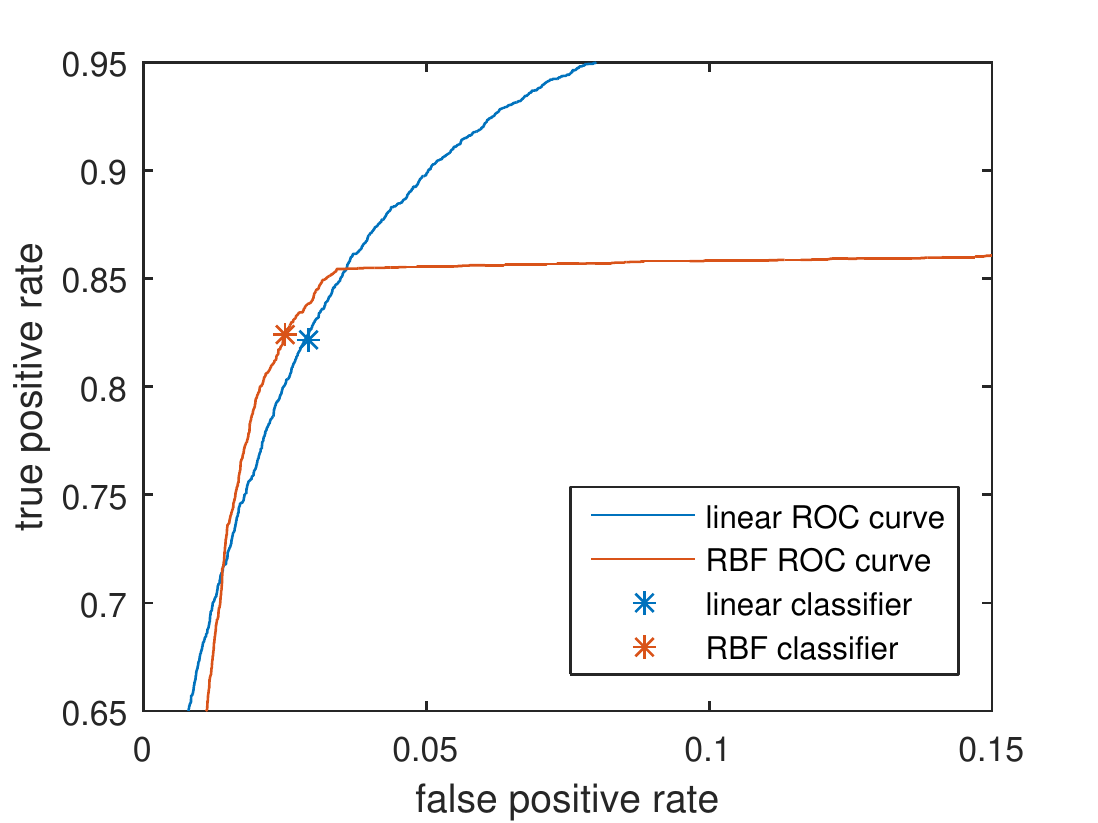}}
\caption{(a) ROC curves for two SVM classifiers using linear and RBF kernels, with specific values for the unbiased  (i.e. $\beta=0$) classifiers. (b) zoom on the upper-left corner.}
\label{roc}
\end{figure}

ROC curves (figure \ref{roc}) give us another way to look at this dilemma between linear and RBF kernels. If we look at the unbiased (i.e. $\beta=0$) classifiers, the RBF is clearly superior: it has fewer false positives and slightly fewer false negatives than the linear classifier. However, when we choose a negative limit $\beta$ (see equation \ref{beta}), for example $\beta = -0.5$, then some of the weakest signals end up over the limit ($f_{1000}^{\text {RBF}}(X) > \beta$) and thus have an estimated label of $\hat{l}(\beta)=1$. Since these weak signals are so common, the false positive rate becomes extremely high.

\subsection{PGA-based (resp. L-based) fragility curve}

In the previous section we always used the score $f_n(X)$ as the parameter on the x-axis to build the fragility curves. However, our method assigns a probability $p(X)$ to each signal, depending only on a few parameters. If we consider this probability as a function of 4 parameters ($p(L,V,PGA,\omega_0)$ if $X \in \R^4$), then we can use any of those parameters, the PGA for example, to define \textit{a posteriori} a fragility curve depending on just this parameter, averaging over the other ones:

\begin{equation} p(PGA) = \mathbb{E} [ p(X) \vert PGA ] . \end{equation}

\begin{figure}[!ht]
\centering
\subfloat[\label{z(f)a}]{\includegraphics[width=6.5cm]{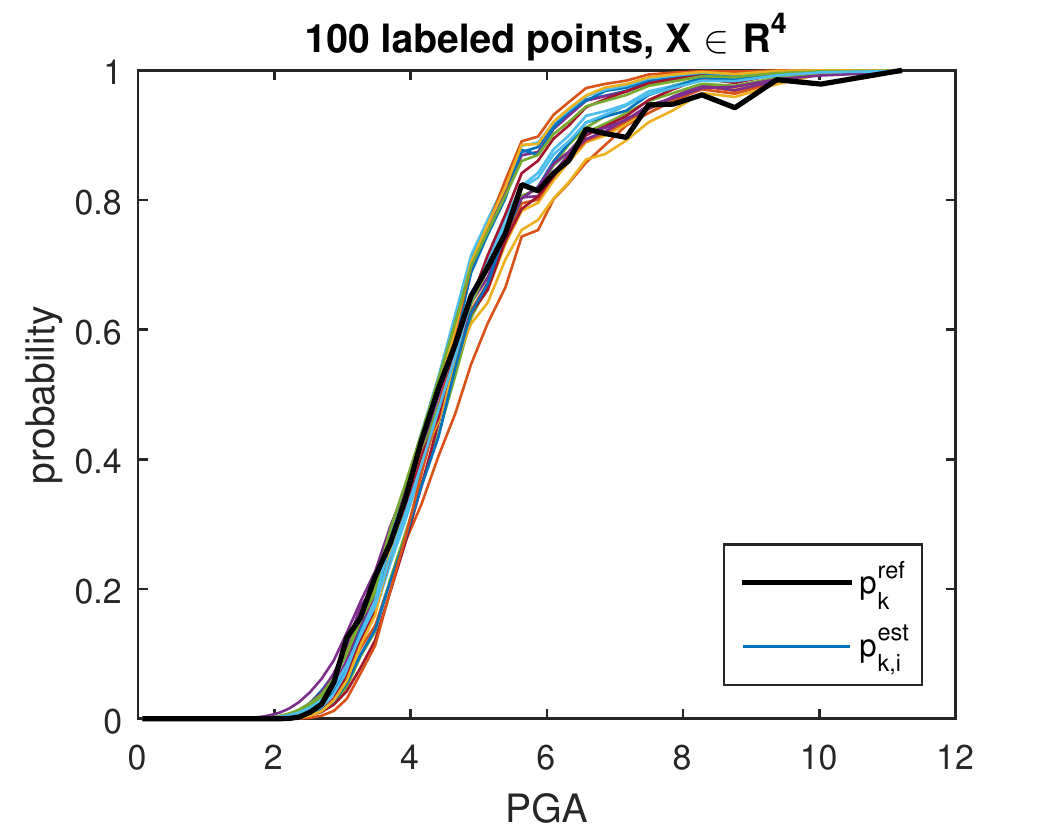}}
\subfloat[\label{z(f)a}]{\includegraphics[width=6.5cm]{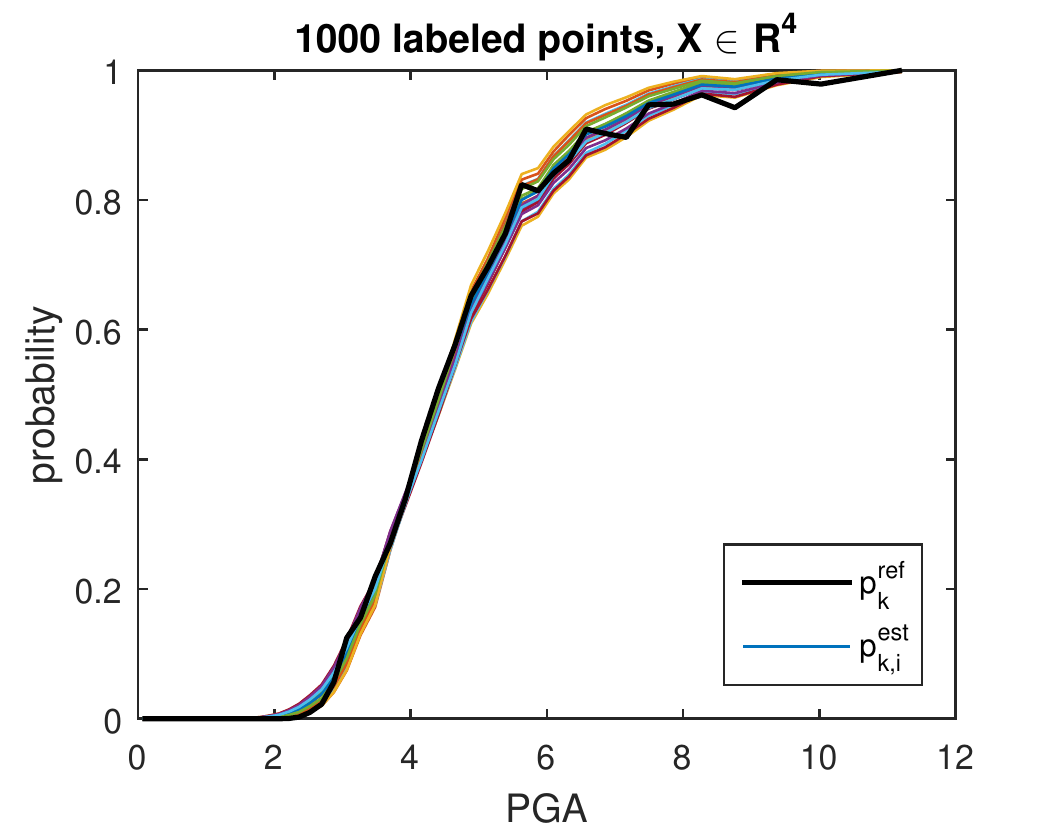}}
\caption{Reference and estimated fragility curves as a function of the PGA, using (a) 100 and (b) 1000 labeled points.}
\label{pga}
\end{figure}

\begin{figure}[!ht]
\centering
\subfloat[\label{z(f)a}]{\includegraphics[width=6.5cm]{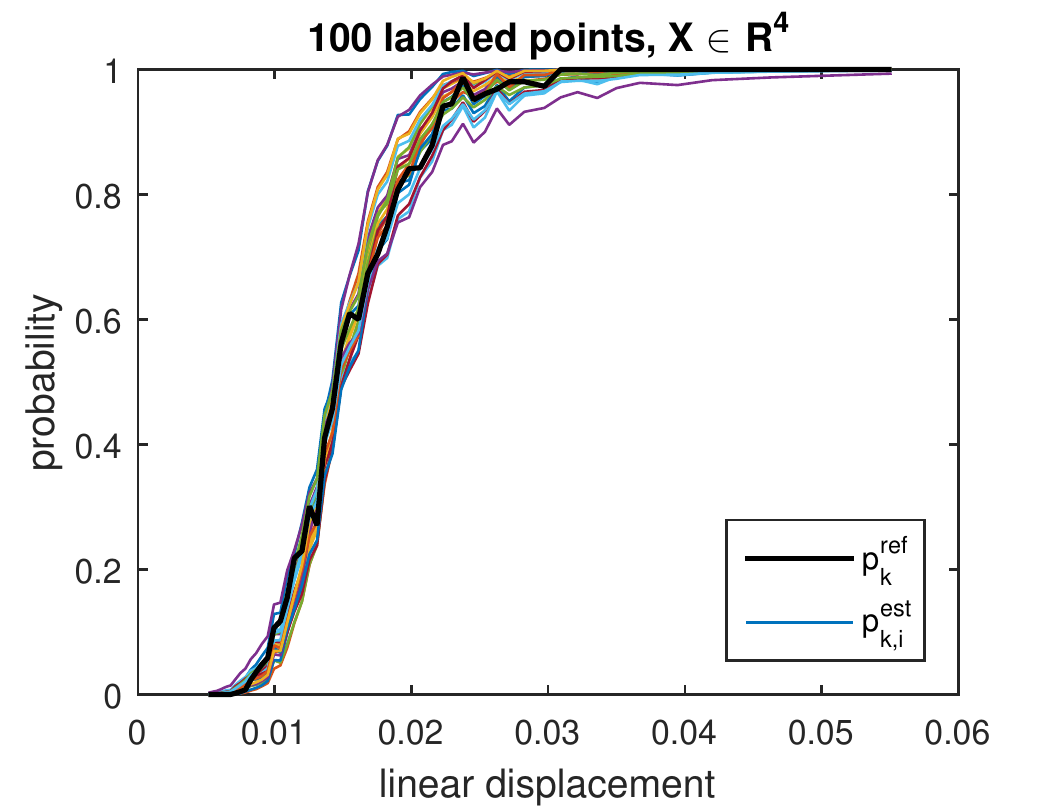}}
\subfloat[\label{z(f)a}]{\includegraphics[width=6.5cm]{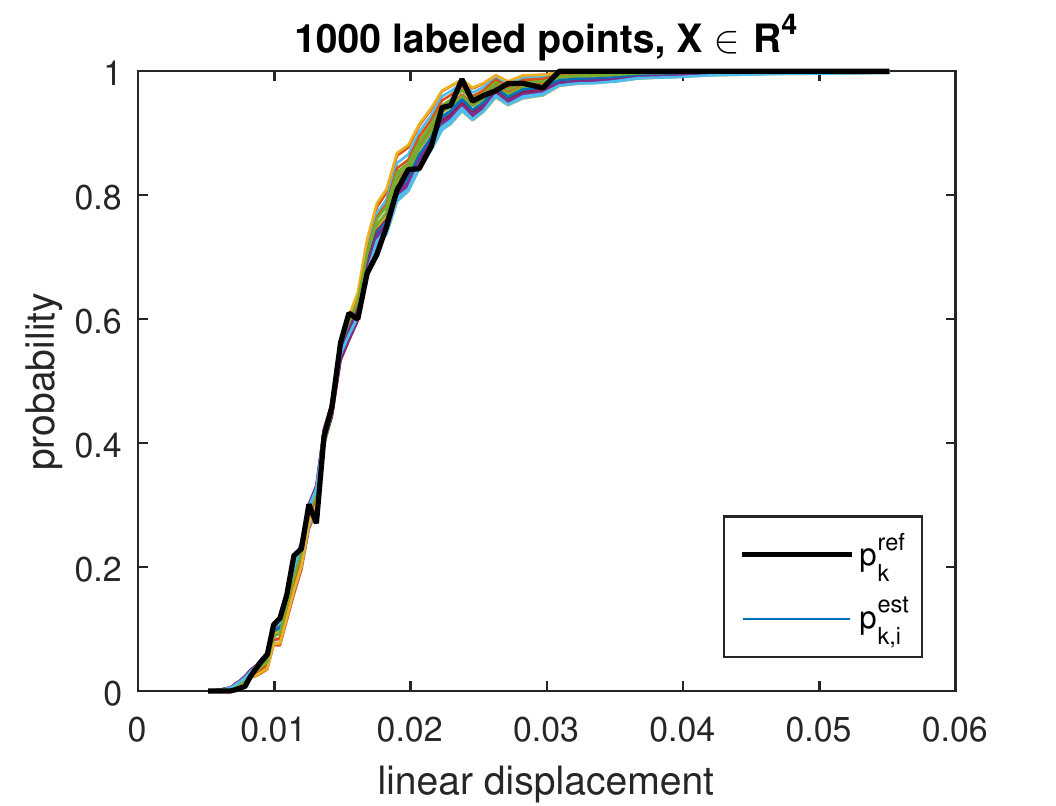}}
\caption{Reference and estimated fragility curves as a function of $L$, using (a) 100 and (b) 1000 labeled points.}
\label{lin}
\end{figure}

Figures \ref{pga} and \ref{lin} show two examples of such curves, using the PGA or the maximum linear displacement $L$. We used a linear SVM classifier in $\R^4$ with 100 and 1000 iterations and computed the probabilities $p(X)$ as before, but then divided the database in groups (with k-means) depending on their PGA (resp. on $L$), instead of the score, before computing the reference probabilities $p_k^{ref}$ and estimated probabilities $p_k^{est}$ for each group $k$. In this case, we can show all 20 test cases in a single figure, since they share a common x-axis (which is not true when we used the score).

\newpage

%The estimated curve is not a logistic function any more. The PGA fragility curve actually looks like a lognormal distribution function, which can be explained by the fact that our physical model is very simple (although our methodology can be applied to any complex structure, in which case the fragility curve might not look like a lognormal function \cite{Mai2017}).

We now have a fully non-parametric fragility curve. The distance between the reference and estimated curve is very small in both cases, even using just 100 labeled instances, although the spread is smaller when we add more data points.

%\subsubsection*{Trading precision for steepness}
\subsection{Trading precision for steepness}

The PGA-based and L-based fragility curves (figures \ref{pga} and \ref{lin}) are very close to the reference curves; the distance $\Delta_{L_2}$ between reference and estimated curves is very small, even smaller than in the case of score-based fragility curves. In this case, why even bother with score-based fragility curves ? What is their benefit, compared to easily-understandable, commonly accepted PGA-based curves ?

The difference is in the steepness of the curve. Formally, when we construct a fragility curve, we choose a projection $F: \R^4 \mapsto \R$ to use as the x-axis. This projection $F(X)$ can be one of the 4 variables (for example the PGA), or the score $f_n(X)$, which can be a linear or nonlinear (in the case of RBF kernel) combinaison of the 4 variables. We then use the k-means algorithm to make groups of signals who are "close" according to this projection, i.e. signals with  the same PGA or the same score; then we compute the estimated probability $p_k^{est}$ for each group. Let us assume for a while that our estimation is very precise, so that $p_k^{est} = p_k^{ref}$ $\forall k$. In this case, which fragility curve gives us the most information ? To see this, we define:
\begin{equation} R^{(F)} = \frac{1}{N} \sum_{k=1}^K n_k \phi (p_k^{est (F)}). \end{equation}
for some nonnegative-valued function $\phi$.
Intuitively, a perfect classifier would give each signal a probability of $0$ or $1$, while a classifier which assigns a probability of $1/2$ to many signals is not very useful. Therefore, we want $\phi$ to be positive on $(0,1)$, 
equal to $0$ for $p=0$ and $p=1$. If we choose:

\begin{equation} \phi(p) = -p \ln (p), \end{equation}
then $R^{(F)}$ can be seen as the entropy of the probability $p(X)$, which would be equal to $0$ for a perfect classifier and has higher values for a "useless" classifier. Another choice would be:

\begin{equation} \phi(p) = \mathbb{1}_{p \in [0.1, 0.9]}. \label{indicatrice}\end{equation}

In this case, $R^{(F)}$ also has a clear physical meaning: it is the proportion of "uncertain" signals, i.e. signals such that $p_k^{est}(X) \in [0.1, 0.9]$. Table \ref{entropy} shows the value of $R^{(F)}$, using the entropy version, for different choices of projection (score, PGA, or linear displacement). We can see on this table that the PGA- and L-based fragility curves are extremely precise, with very low values of $\Delta_{L_2}$ (this can also be seen in figures \ref{pga} and \ref{lin}), but their entropy is much higher than the score-based fragility curves.

\begin{table}[!h]
\centering
  \begin{tabular}{| c | c | c | c | c |}
   \hline
   & projection & score & PGA & L \\ \hhline{|=|=|=|=|=|}
   n=100 & $\Delta_{L_2}$ ($\%$) & $3.8 \pm 1.6$ & $2.6 \pm 1$ & $2.8 \pm 0.9$ \\ \cline{2-5}
   & entropy $(10^{-2})$ & $5.3 \pm 1.7$ & $12.3 \pm 1.8$ & $12.2 \pm 2$ \\ \hhline{|=|=|=|=|=|}
   n=1000 & $\Delta_{L_2}$ ($\%$) & $1.7 \pm 0.6$ & $1.6 \pm 0.3$ & $1.4 \pm 0.4$ \\ \cline{2-5}
   & entropy $(10^{-2})$ & $7.2 \pm 1.2$ & $13.3 \pm 1.5$ & $13.6 \pm 1$ \\ \hline
  \end{tabular}
\caption{Precision and entropy of fragility curves using different projections (average and standard deviation over 20 test cases), for n=100 (top) or n=1000 (bottom) labeled points.}\label{entropy}
\end{table}

One surprising fact of table \ref{entropy} is that the entropy is smaller at $n=100$ compared to $n=1000$ in all three cases. This shows that after only $n=100$ mechanical calculations, all our classifiers tend to slightly overestimate the steepness, and give fragility curves that are actually steeper than the reality (and also steeper than the more realistic $n=1000$ curves). This was also seen in figures \ref{4rbfb} and \ref{4rbfe}: at $n=100$ iterations the estimated curve is steeper than the reference curve, which gives an estimated entropy smaller than the reference entropy. Using the other choice of $\phi$ gives the same conclusions: the proportion of signals with $p_k^{est} \in [0.1,0.9]$ is $18.2\%$ if we use the score, but it is around $28\%$ for both the PGA and maximum linear displacement. Therefore, the choice of the projection used for a fragility curve is a trade between precision and steepness. Keep in mind that the values of the entropy for different choices of projection can be obtained after the active learning, and the computationnal cost is very small (mostly the cost of k-means). As a consequence, this choice can be made a posteriori, from the probabilities assigned to each signal.

%However, these curve gives us less information than the fragility curve based on the score (figure \ref{4rbf} middle right). Both are very precise (the error is small), but the difference is their \textit{steepness}. A perfect classifier would give each signal a probability of $0$ or $1$; a good classifier tries to minimize the number of "uncertain" signals, for example the signals such that $p^{ref}(X) \in [0.1, 0.9]$.\\
%\\
%The PGA fragility curve in figure \ref{pga} gives us $29\%$ of uncertain signals (such that $p^{ref}(X) \in [0.1, 0.9]$), and only $2.5\%$ have $p^{ref}(X) > 0.9$. As a comparison, figure \ref{raideur} shows the steepness of our 3 classifiers as a function of the number of iterations. The linear SVMs in $\R^4$ and $\R^{13}$ give similar results, with $17/18\%$ of uncertain signals after 1000 iterations. The RBF kernel is much steeper, with only $9.3\%$ of uncertain signals. The true proportion of signals that exceeds the threshold in the preprocessed data is around $17\%$.

%\begin{figure}[h]
%\centering
%\includegraphics[scale=0.6]{raideur2.pdf}
%\caption{Proportion of "uncertain" signals with $p^{ref}(X) \in [0.1, 0.9]$, for the PGA fragility curve of figure \ref{pga} (black) and for the 3 active learners.}
%\label{raideur}
%\end{figure}

\subsection{Additionnal remarks}
\subsubsection{About the specificity of active learning}

\begin{figure}[h]
\centering
\includegraphics[width=6.5cm]{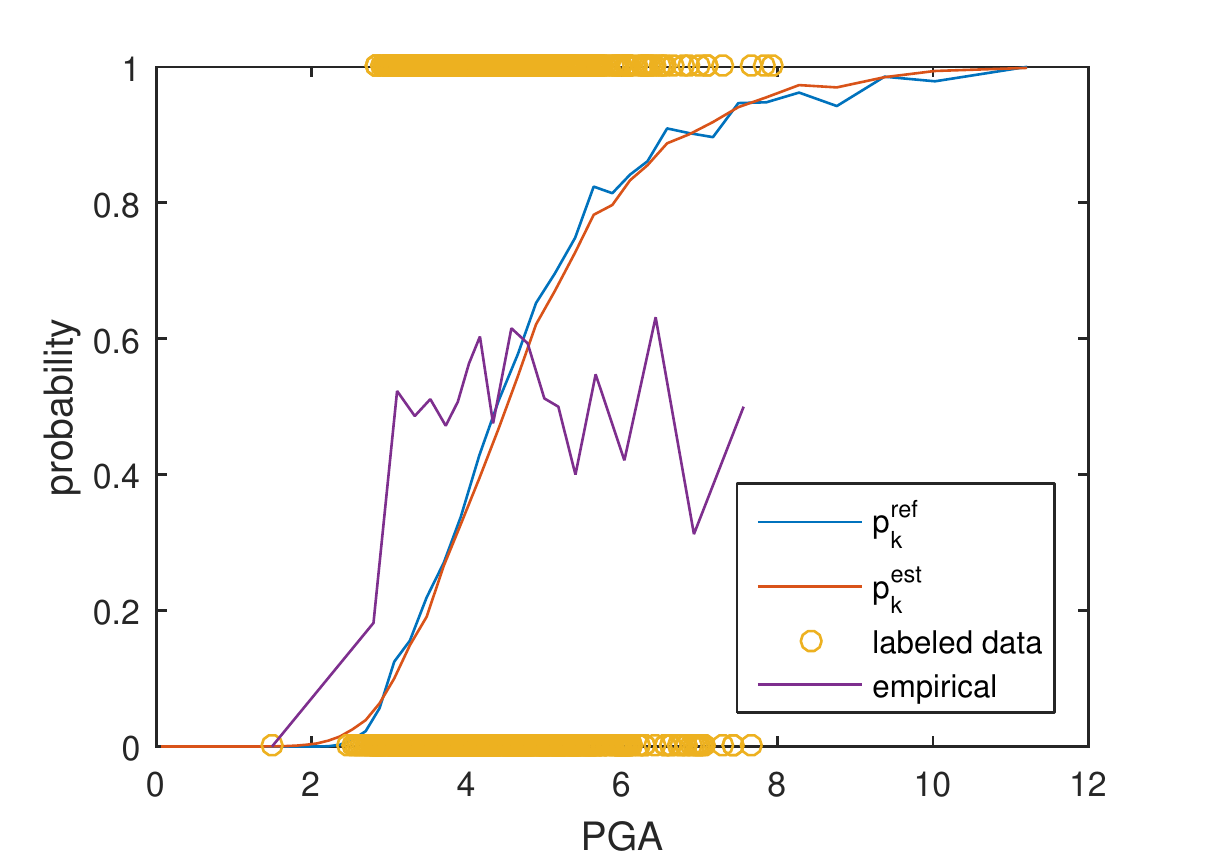}
\caption{Empirical fragility curve using only the labeled set (violet).}
\label{emp}
\end{figure}

Using the score to compute the probabilities (equation \ref{p(x)}) on the whole dataset $\mathcal{X}$ is absolutely mandatory, even if one is only interested in the PGA fragility curves. In particular, looking only at the labeled set $\mathcal{L}$ to find directly a probability of failure depending on the PGA gives extremely wrong results. Figure \ref{emp} shows not only the reference and estimated fragility curves previously defined, but also the $n=1000$ points of the labeled set $\mathcal{L}$ and an empirical probability built from it.

This empirical fragility curve was computed by using the k-means algorithm on the PGA values of the labeled set $\mathcal{L}$, then taking the empirical failure probability in each group. The result looks like... a constant around $0.5$ (with high variability). Looking only at $\mathcal{L}$, the PGA does not even look correlated with failure. The reason for this surprising (and completely wrong) result is the active learning algorithm. $\mathcal{L}$ is not at all a random subset of $\mathcal{X}$; it is a carefully chosen set of the data points with maximal uncertainty, which means that most of them end up very close to the final hyperplane (and have a final score very close to $0$), but also that they span as much of this hyperplane as possible. When these points are projected on any axis not perpendicular to the hyperplane (that is, any axis but the score), for example the PGA (which is one of the components of $X$), the empirical probability is roughly equal to the   constant $1/2$, which is not representative at all of the underlying probability. Using equation \ref{p(x)} (then eventually fitting the results to a lognormal curve for the PGA) is the right way to go.

\subsubsection{About a combination of the linear and RBF kernels}

The RBF kernel was very promising in terms of classification, as we saw on the first results (figure \ref{res}); however, we saw in the following sections that using it to make a fragility curve can lead to catastrophic results (figures \ref{4rbfc} and  \ref{4rbff}). Could we combine the two kernels in a way that let us keep the benefits of both ? One simple way to do that is to use the following procedure:
\begin{itemize}
\item use the active learning with RBF kernel to select $n=1000$ signals to be labeled;
\item from these 1000 data points, train two classifiers, one with a linear kernel, the other with the RBF kernel;
\item assign two scores $f_{lin}(X)$ and $f_{rbf}(X)$ to each non-labeled signal, using the two classifiers;
\item fit the labeled points to each set of scores, giving you two probabilities $p_{lin}(X)$ and $p_{rbf}(X)$ for every signal;
\item the "final" probability is chosen as:

\begin{equation} p(X) = \left\lbrace \begin{array}{rl} p_{lin}(X) \quad & \text{if} \quad p_{lin}(X) < 0.05 \text{ or } p_{lin}(X) > 0.95, \\ p_{rbf}(X) \quad & \text{otherwise}. \end{array} \right. \end{equation}

\end{itemize}

Since the active learning used RBF kernels and we use the RBF score for any "uncertain" signal, the PRBP has about the same value than in the "pure RBF" version, around 0.85. However, the $\Delta_{L_2}$ is at $2.3\%$, slightly higher than for the "pure linear" version (1.8\%), but much better than the catastrophic "pure RBF" version (20\% !). Although this procedure may seem to have the best of both worlds, in a practical application the PRBP may not be very interesting if the goal is to make a fragility curve; in this case only the precision and steepness of the curve are important, and a linear kernel performs better than a RBF kernel.

\section{Conclusion}

This paper proposed an efficient methodology for estimating non-parametric seismic fragility curves by active learning with a Support Vector Machines classifier. We have introduced and studied this methodology when aleatory uncertainties have a predominant contribution in the variability of structural response, that is to say when the contribution of uncertainties regarding seismic excitation is "much larger" than the contribution of uncertainties regarding structural capacity. In this work, structure was considered as deterministic.
In this framework, a perfect classifier, if it exists, would lead to a fragility curve in the form of a unit step function, i.e. corresponding to a fragility curve "without uncertainty". That means the output of this classifier, which is a score, would be the best seismic intensity measure indicator to evaluate the damaging potential of the seismic signals, knowing that such a classifier would necessary be both structure and failure criterion-dependent, with possibly a dependence on the ground motion characteristics (near-fault type like, broadband, etc).

The proposed methodology makes it possible to build such a (non-perfect) classifier. It consists in (i) reducing input excitation to some relevant parameters and, given these parameters, (ii) using a SVM for a binary classification of the structural responses relative to a limit threshold of exceedance. Selection of the mechanical numerical calculations by active learning dramatically reduces the computational cost of construction of the classifier. The output of the classifier, the score, is the desired intensity measure indicator which is then interpreted in a probabilistic way to estimate fragility curves as score functions or as functions of classical seismic intensity measures.

This work shows that a simple but crucial preprocessing of the data (i.e. Box-Cox transformation of the input parameters) makes it possible to use a simple linear SVM to obtain a very precise classifier after just one hundred iterations, that is to say with one hundred mechanical calculations. Moreover, for the class of structures considered, with only four classical seismic parameters ($PGA$, $V$, $L$, $\omega_0$), the score-based fragility curve is very close to the reference curve (obtained with a massive Monte Carlo-based approach) and steeper than the PGA-based one, as expected. L-based fragility curves appear to perform about as well as PGA-based ones in our setting. Advanced SVMs using RBF kernel result in less classification errors when using one thousand mechanical calculations, but does not appear well suited to making fragility curves.

A naive way to take into account epistemic uncertainties would consist in building a classifier for each set of structural parameters. Nevertheless, such a method would not be numerically efficient. Another way could be to assume that epistemic uncertainties have small influence on the classifier evaluated, for example, for the median capacity of the structure. Thus, only calculations of linear displacements would be necessary to estimate the corresponding fragility curve. However, to avoid such assumptions, some research efforts have to be devoted to propose an efficient overall methodology that takes into account the two types of uncertainties.

%\nocite{*}% Show all bib entries - both cited and uncited; comment this line to view only cited bib entries;
\bibliographystyle{abbrv}
\bibliography{biblio}%

\end{document}